\documentclass[10pt,journal,compsoc]{IEEEtran}

\ifCLASSINFOpdf
\else
\usepackage[dvips]{graphicx}
\fi
\usepackage{url}

\usepackage[colorlinks,linkcolor=blue,anchorcolor=blue,
citecolor=blue]{hyperref}
\usepackage{graphicx}
\usepackage{amsmath,amssymb}
\usepackage{booktabs,tabularx,multirow,threeparttable}
\usepackage{float}
\usepackage{framed}
\usepackage{subfigure}
\usepackage{amssymb}
\usepackage{threeparttable}
\usepackage{subfigure}
\ifCLASSOPTIONcompsoc
  \usepackage[nocompress]{cite}
\else
  \usepackage{cite}
\fi

\begin{document}
%\title{InfDet: Instance-Level Feature Denoising for Robust Aerial Image Detection}
%\title{InfDet: Towards More Robust Detection for Small, Cluttered and Rotated Objects via Instance-Level Feature Denoising}
\title{SCRDet++: Detecting Small, Cluttered and Rotated Objects via Instance-Level Feature Denoising and Rotation Loss Smoothing}
%in Aerial Images
%~\IEEEmembership{Member,~IEEE,}
\author{Xue~Yang, Junchi~Yan~\IEEEmembership{Senior Member,~IEEE}, Wenlong Liao, \\Xiaokang Yang~\IEEEmembership{Fellow,~IEEE}, Jin Tang, Tao He
%, Xiaokang~Yang~\IEEEmembership{Fellow,~IEEE}
%Yingxin~Lou, Tao~Xue,	
%\thanks{Manuscript received April 19, 2005; revised August 26, 2015. \textit{(Corresponding author: Junchi~Yan)}
\thanks{X.~Yang, J.~Yan, W. Liao, X. Yang are with Department of Computer Science and Engineering, Shanghai Jiao Tong University, and also with the MoE Key Lab of Artificial Intelligence, AI Institute, Shanghai Jiao Tong University. Jin Tang is with Anhui Province Key Laboratory of Multimodal Cognitive Computation, and School of Computer Science and Technology, Anhui University. Tao He is with COWAROBOT Co., Ltd., and Anhui Province Key Laboratory of Multimodal Cognitive Computation.}
\thanks{Correspondence author: Junchi Yan}
\thanks{E-mail: \{yangxue-2019-sjtu,yanjunchi,igoliao,xkyang\}@sjtu.edu.cn, tj@ahu.edu.cn, tommie.he@cowarobot.com}
}
\markboth{}%
{Shell \MakeLowercase{\textit{et al.}}: Bare Demo of IEEEtran.cls for IEEE Journals}

%Though considerable progress has been made, there still exist challenges for objects in complex scenes. Apart from natural images, such issues are especially pronounced for aerial images.
%We show our technique can boost the performance of horizontal detection and rotation detection in remote sensing imagery.
\IEEEtitleabstractindextext{
\begin{abstract}
Small and cluttered objects are common in real-world which are challenging for detection. The difficulty is further pronounced when the objects are rotated, as traditional detectors often routinely locate the objects in horizontal bounding box such that the region of interest is contaminated with background or nearby interleaved objects. In this paper, we first innovatively introduce the idea of denoising to object detection. Instance-level denoising on the feature map is performed to enhance the detection to small and cluttered objects. To handle the rotation variation, we also add a novel IoU constant factor to the smooth L1 loss to address the long standing boundary problem, which to our analysis, is mainly caused by the periodicity of angular (PoA) and exchangeability of edges (EoE). By combing these two features, our proposed detector is termed as SCRDet++. Extensive experiments are performed on large aerial images public datasets DOTA, DIOR, UCAS-AOD as well as natural image dataset COCO, scene text dataset ICDAR2015, small traffic light dataset BSTLD and our released S$^2$TLD by this paper. The results show the effectiveness of our approach. The released dataset S$^2$TLD is made public available, which contains 5,786 images with 14,130 traffic light instances across five categories.
\end{abstract}

}

\maketitle
\section{Introduction}

\IEEEPARstart{O}{bject} detection is one of the fundamental tasks in computer vision and various general-purpose detectors~\cite{girshick2014rich,he2014spatial,girshick2015fast,liu2016ssd,redmon2016you,dai2016r,ren2017faster} based on convolutional neural networks (CNNs) have been devised. Promising results have been achieved on public benchmarks including MS COCO \cite{lin2014microsoft} and VOC2007 \cite{everingham2010pascal} etc. However, most existing detectors do not pay particular attention to some common aspects for robust object detection in the wild: small size, cluttered arrangement and arbitrary orientations. These challenges are especially pronounced for aerial image~\cite{xia2018dota,li2020object, cheng2016learning, liu2017high} which has become an important area for detection in practice, for its various civil applications, e.g. resource detection, environmental monitoring, and urban planning. 

%which is not only helpful in military confrontation to obtain precise battlefield information (e.g. strategic targets and accurate positioning), but also plays a vital role in civil applications e.g. resource detection, environmental monitoring, and urban planning. 

\begin{figure}[!tb]
	\centering
	\subfigure[Horizontal detection.]{
		\begin{minipage}[t]{0.47\linewidth}
			\centering
			\includegraphics[width=1.0\textwidth]{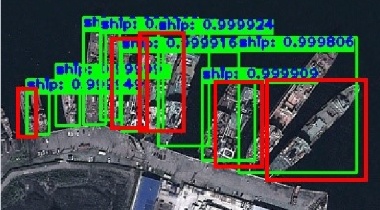}
			\centering
			\label{fig:ship_h}
	\end{minipage}}
	\subfigure[Rotation detection.]{
		\begin{minipage}[t]{0.46\linewidth}
			\centering
			\includegraphics[width=1.0\textwidth]{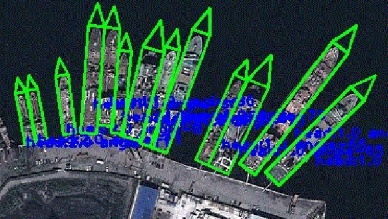}
			\centering
			\label{fig:ship_r}
	\end{minipage}}
	\caption{Small, cluttered and rotated objects in complex scene whereby rotation detection plays an important role. Red boxes indicate missing detection which are suppressed by non-maximum suppression (NMS).}
	\label{fig:ship}
	%\vspace{-10pt}
\end{figure}

In the context of remote sensing, we also present some specific discussion to motivate this paper, as shown in Fig. \ref{fig:ship}. It shall be noted that these three aspects also prevail in other scenarios e.g. natural images and scene texts.

1) \textbf{Small objects.} Aerial images often contain small objects overwhelmed by complex surrounding scenes.

2) \textbf{Cluttered arrangement.} Objects e.g. vehicles and ships in aerial images are often densely arranged, leading to inter-class feature coupling and intra-class feature boundary blur.

3) \textbf{Arbitrary orientations.} Objects in aerial images can appear in various orientations. Rotation detection is necessary especially considering the high aspect ratio issue: the horizontal bounding box for a rotated object is more loose than an aligned rotated one, such that the box contains a large portion of background or nearby cluttered objects as disturbance. Moreover, it will be greatly affected by non-maximum suppression, see Fig. \ref{fig:ship_h}.
  
As described above, the small/cluttered objects problem can be interleaved with the rotation variance. In this paper, we aim to address the first challenge by seeking a new way of dismissing the noisy interference from both background and other foreground objects. While for rotation alignment, a new rotation loss is devised accordingly. Our both techniques can serve as plug in for existing detectors~\cite{ren2017faster, lin2017feature, lin2017focal, ma2018arbitrary, jiang2017r2cnn,  yang2021r3det}, in an out of box manner. We give further description as follows.

%~\cite{zhang2017beyond,jia2019focnet,chen2018image, brooks2019unprocessing,yang2017estimation,godard2018deep,zhang2018learning, kokkinos2019iterative}
For small and cluttered object detection, we devise a denoising module and in fact denoising has not been studied for objection detection. We observe two common types of noises that are orthogonal to each other: i) image level noise, which is object-agnostic, and ii) instance level noise, specifically often in the form of mutual interference between objects, as well as background interference. Such noises are ubiquitous and pronounced in aerial images which are remotely sensed. In fact, denoising has been a long standing task~\cite{tian2020deep,xie2019feature,milani2012adaptive,cho2020dapas} in image processing while they are rarely designated for object detection, and the denoising is finally performed on raw image for the purpose of image enhancement rather than downstream semantic tasks, especially in an end-to-end manner.

In this paper, we explore the way of performing instance level denoising (InLD) and particularly in the feature map (i.e. latent layers' outputs by CNNs), for robust detection. The hope is to reduce the inter-class feature coupling and intra-class interference, meanwhile blocking background interference. To this end, a novel InLD component is designated to decouple the features of different object categories into their respective channels approximately. Meanwhile, in the spatial domain, the features of the object and background are enhanced and weakened, respectively. It is worth noting that the above idea is conceptually similar to but inherently different from the recent efforts~\cite{xie2019feature, cho2020dapas} for image level feature map denoising (ImLD), which is used as a way of enhancing the image recognition model's robustness against attack, rather than location sensitive object detection. Readers are referred to Tab. \ref{table:ImLD_and_InLD} for a quick verification that our InLD can more effectively improve detection than ImLD for both horizontal and rotation cases.

On the other hand, as discussed above, as a closely interleaved problem to small/cluttered object detection, accurate rotation estimation is addressed by devising a novel IoU-Smooth L1 loss. It is motivated by the fact that the existing state-of-the-art regression-based rotation detection methods e.g. five-parameter regression~\cite{azimi2018towards, ding2018learning,  yang2021r3det, zhang2019cad} suffer from the issue of discontinuous boundaries, which is inherently caused by the periodicity of angular (PoA) and exchangeability of edges (EoE) \cite{yang2020arbitrary} (see details in Sec.~\ref{sec:rod}).

%Although region regression-based rotation detection methods have achieved promising results, there are still some fundamental problems. We discover the five-parameter regression methods have the problem of  Therefore, we propose IoU-Smooth L1 loss to address the boundary problem in rotation detection 

We conduct extensive ablation study and experiments on multiple datasets including both aerial images from DOTA \cite{xia2018dota}, DIOR \cite{li2020object}, UCAS-AOD \cite{li2019feature}, as well as natural image dataset COCO \cite{lin2014microsoft}, scene text dataset ICDAR2015 \cite{karatzas2015icdar}, small traffic light dataset BSTLD \cite{behrendt2017deep} and our newly released S$^2$TLD to illustrate the promising effects of our techniques. 

%nd demonstrate that our method achieves the state-of-the-art detection accuracy in both rotation detection and horizontal detection. We also use  to further verify the effectiveness of the proposed techniques.
%i) adoption and verification of the image denoising module to aerial image detection, which has not been studied in literature before; i
%integrating our instance-level feature denoising module into other detection models with thorough ablation study.

\begin{figure*}[!tb]
	\begin{center}
		\includegraphics[width=0.85\linewidth]{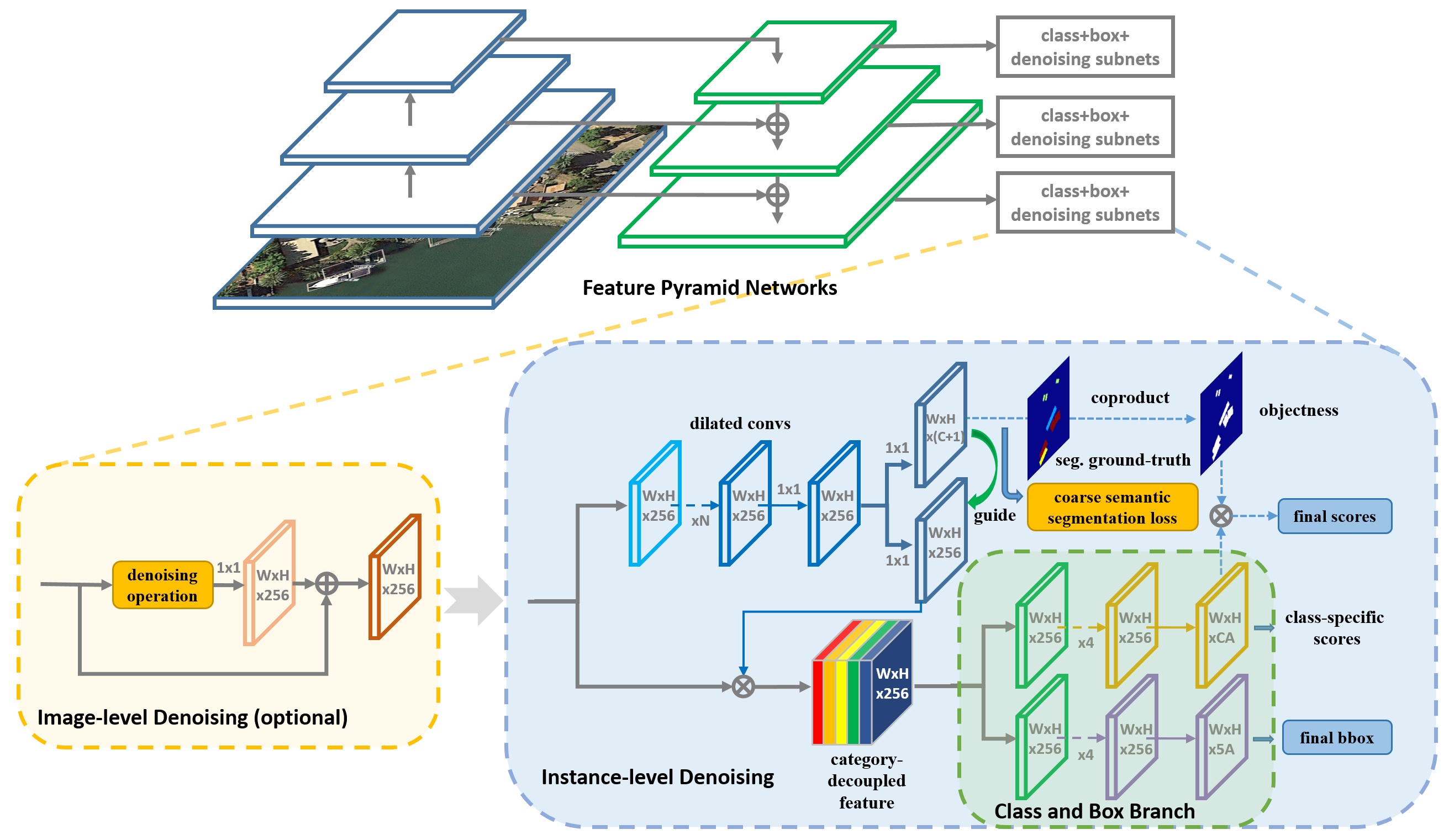}
	\end{center}
	%\vspace{-8pt}
	\caption{The pipeline of our method (using RetinaNet~\cite{lin2017focal} as an embodiment). Our SCRDet++ mainly consists of four modules: basic embodiment for feature extraction, Image-level denoising for removing common image noise, instance-level denoising module for suppressing instance noise (i.e., inter-class feature coupling and distraction between intra-class and background) and the `class+box' branch for predicting classification score and bounding box position. `C' and `A' represent the number of object categories and the number of anchor at each feature point, respectively.}
	\label{fig:pipeline}
\end{figure*}

The preliminary content of this paper has partially appeared in the conference version~\cite{yang2019scrdet}\footnote{Compared with the conference version, this journal version has made the following extensions: i) we take a novel feature map denoising perspective to the small and cluttered object detection problem, and specifically devise a new instance-level feature denoising technique for detecting small and cluttered objects with little additional computation and parameter overhead; ii) comprehensive ablation study of our instance-level feature denoising component across datasets, which can be easily plugged into existing detectors. Our new method significantly outperforms our previous detector in the conference version (e.g. overall detection accuracy 72.61\% versus 76.81\%, and 75.35\% versus 79.35\% on the OBB and HBB task of DOTA-v1.0 dataset, respectively); iii) We collect, annotate and release a new small traffic light dataset (5,786 images with 1,4130 traffic light instances across five categories) to further verify the versatility and generalization performance of the instance-level denoising module; iv) last but not least, the paper has been largely rephrased and expanded to cover the discussion of up-to-date works including those on image denoising and small object detection. The source code is also released.}, with the detector named SCRDet (Small, Cluttered, and Rotated Object Detector). In this journal version, we extend our improved detector called SCRDet++. The overall contributions are:

%1) The image denoising module is adopted and positively verified for aerial image detection, whereby the objects are often small and densely distributed. This has not been studied in literature before.

%, which is shown can effectively reduce the inter-class feature coupling and intra-class feature interference, while blocking interference from non-object area
1) To our best knowledge, we are the first to develop the concept of instance level noise (at least in the context of object detection), and design a novel Instance-Level Denoising (InLD) module in feature map. This is realized by supervised segmentation whose ground truth is approximately obtained by the bounding box in object detection. The proposed module effectively addresses the challenges in detecting small size, arbitrary direction, and dense distribution objects with little computation and parameter increase.

2) Towards more robust handling of arbitrarily-rotated objects, an improved smooth L1 loss is devised by adding the IoU constant factor, which is tailored to solve the boundary problem of the rotating bounding box regression.

3) We create and release a real-world traffic light dataset: S$^2$TLD. It consists of 5,786 images with 14,130 traffic light instances across five categories: red, green, yellow, off and wait on. It further verifies the effectiveness of InLD, and it is available at \url{https://github.com/Thinklab-SJTU/S2TLD}.

4) Our method achieves state-of-the-art performance on public datasets for rotation detection in complex scenes like the aerial images. Experiments also show that our InLD module, which can be easily plugged into existing architectures, can notably improve detection on different tasks.

%This paper is organized as follows. Related work is presented and discussed in Section~\ref{sec:related}. Section~\ref{sec:method} presents the proposed approach and the experiments are shown in Section~\ref{sec:experiment}. Section~\ref{sec:conclusion} concludes this paper.

\section{Related Work}\label{sec:related}
We first discuss existing detectors for both horizontal bounding box based detection and rotation detection. Then some representative works on image denoising and small object detection are also introduced. 
\subsection{Horizontal Region Object Detection}
There is an emerging line of deep network based object detectors. R-CNN \cite{girshick2014rich} pioneers the CNN-based detection pipeline. Subsequently, region-based models such as Fast R-CNN \cite{girshick2015fast}, Faster R-CNN \cite{ren2017faster}, and R-FCN \cite{dai2016r} are proposed, which achieves more cost-effective detection. SSD \cite{liu2016ssd}, YOLO \cite{redmon2016you} and RetinaNet \cite{lin2017focal} are representative single-stage methods, and their single-stage structure further improves detection speed. In addition to anchor-based methods, many anchor-free also have become popular in recent years. FCOS \cite{tian2019fcos}, CornerNet \cite{law2018cornernet}, CenterNet \cite{duan2019centernet} and ExtremeNet \cite{zhou2019bottom} attempt to predict some keypoints of objects such as corners or extreme points, which are then grouped into bounding boxes, and these detectors have also been applied to the field of remote sensing~\cite{wei2020oriented,xiao2020axis}. R-P-Faster R-CNN \cite{han2017efficient} achieves satisfactory performance in small datasets. The method \cite{xu2017deformable} combines both deformable convolution layers \cite{dai2017deformable} and region-based fully convolutional networks (R-FCN) to improve detection accuracy further. The work \cite{ren2018deformable} adopts top-down and skipped connections to produce a single high-level feature map of a fine resolution, improving the performance of the deformable Faster R-CNN model. IoU-Adaptive R-CNN \cite{yan2019iou} reduces the loss of small object information by a new IoU-guided detection network. FMSSD \cite{wang2019fmssd} aggregates the context information both in multiple scales and the same scale feature maps. However, objects in aerial images with small size, cluttered distribution and arbitrary rotation are still challenging, especially for horizontal region detection methods.

\subsection{Arbitrary-Oriented Object Detection}
The demand for rotation detection has been increasing recently like for aerial images and scene texts. Recent advances are mainly driven by the adoption of rotated bounding boxes or quadrangles to represent multi-oriented objects. For scene text detection, RRPN \cite{ma2018arbitrary} employs rotated RPN to generate rotated proposals and further perform rotated bounding box regression. TextBoxes++ \cite{liao2018textboxes++} adopts vertex regression on SSD. RRD \cite{liao2018rotation}  further improves TextBoxes++ by decoupling classification and bounding box regression on rotation-invariant and rotation sensitive features, respectively. EAST \cite{zhou2017east} directly predicts words or text lines of arbitrary orientations and quadrilateral shapes in full images, eliminating unnecessary intermediate steps with a single neural network. Recent text spotting methods like FOTS \cite{liu2018fots} show that training text detection and recognition simultaneously can greatly boost detection performance. In contrast, aerial images object detection is more challenging: first, multi-category object detection requires the generalization of the detector. Second, small objects in aerial images are usually densely arranged on a large scale. Third, aerial image  detection requires a more robust algorithm due to the variety of noises. Many aerial images rotation detection algorithms are designed for different problems. ICN \cite{azimi2018towards}, ROI Transformer~\cite{ding2018learning}, and SCRDet~\cite{yang2019scrdet} are representative of two-stage aerial images rotation detectors, which are mainly designed from the perspective of feature extraction. From the results, they have achieved good performance in small or dense object detection. Compared to the previous methods, R$^3$Det~\cite{ yang2021r3det} and RSDet~\cite{qian2021learning} are based on a single-stage detection method which pay more attention to the trade-off of accuracy and speed. Gliding Vertex \cite{xu2020gliding} and RSDet~\cite{qian2021learning} achieve more accurate object detection via quadrilateral regression prediction. Axis Learning \cite{xiao2020axis} and O$^2$-DNet \cite{wei2020oriented} are combined with the latest popular anchor-free ideas, to overcome the problem of too many anchors in anchor-based detection methods.

%1) additive white noisy images \cite{zhang2017beyond,jia2019focnet}; 2) real noisy images \cite{chen2018image, brooks2019unprocessing}; 3) blind denoising \cite{yang2017estimation, godard2018deep} and 4) hybrid noisy images \cite{zhang2018learning, kokkinos2019iterative}
\subsection{Image Denoising}
Deep learning has obtained much attention in image denoising. The survey \cite{tian2020deep} divides image denoising using CNNs into four types (see the references therein): 1) additive white noisy images; 2) real noisy images; 3) blind denoising and 4) hybrid noisy images, as the combination of noisy, blurred and low-resolution images. In addition, image denoising also helps to improve the performance of other computer vision tasks, such as image classification \cite{xie2019feature}, object detection \cite{milani2012adaptive}, semantic segmentation \cite{cho2020dapas}, etc. In addition to image noise, we find that there is also instance noise in the field of object detection. Instance noise describes object-aware noise, which is more widespread in object detection than object-agnostic image noise. In this paper, we will explore the application of image-level denoising and instance-level denoising techniques to object detection in complex scenes.

\subsection{Small Object Detection}
Small object detection remains an unsolved challenge. Common small object solutions include data augmentation \cite{kisantal2019augmentation}, multi-scale feature fusion \cite{lin2017feature, deng2021extended}, tailored sampling strategies \cite{zhu2018seeing, liu2020hambox, yang2019scrdet}, generative adversarial networks \cite{li2017perceptual}, and multi-scale training \cite{singh2018sniper} etc. In this paper, we show that denoising is also an effective means to improve the detection performance of small objects. In complex scenes, the feature information of small objects is often overwhelmed by the background area, which often contains a large number of similar objects. Unlike ordinary image-level denoising, we will use instance-level denoising to improve the detection capabilities of small objects, which is a new perspective.

This paper mainly considers designing a general-purpose instance level feature denoising module, to boost the performance of horizontal detection and rotation detection in challenging aerial imagery, as well as natural images and scene texts. Besides, we also design an IoU-Smooth L1 loss to solve the boundary problem of the arbitrary-oriented object detection for more accurate rotation estimation.

\section{The Proposed Method}\label{sec:method}
\subsection{Approach Overview}
Fig. \ref{fig:pipeline} illustrates the pipeline of the proposed SCRDet++. It mainly consists of four modules: i) feature extraction via CNNs which can take different forms of CNNs from existing detectors e.g.~\cite{girshick2014rich,liu2016ssd}, ii) image-level denoising (ImLD) module for removing common image noise, which is optional as its effect can be well offset by the subsequent InLD as devised in this paper; iii) instance-level denoising (InLD) module for suppressing instance noise (i.e., inter-class feature coupling and distraction between intra-class and background) and iv) the class and box branch for predicting score and (rotated) bounding box. Specifically, we first describe our main technique i.e. instance-level denoising module (InLD) in Sec. \ref{subsec:InLD}, which further contains a comparison with the image level denoising module (ImLD). Finally, we detail the network learning which involves a specially designed smooth loss for rotation estimation in Sec. \ref{subsec:learning}. Note that in experiments we show that InLD can replace and strike a more effective role for detection than ImLD, making ImLD a dispensable component in our pipeline.

\subsection{Instance-level Feature Map Denoising}\label{subsec:InLD}
In this subsection, we present our devised instance-level feature map denoising approach. To emphasis the importance of our instance-level operation, we further compare it with image-level denoising in feature map, which is also adopted for robust image recognition model learning in \cite{xie2019feature}. To our best knowledge, our approach is the first for using (instance level) feature map denoising for object detection. The denoising module can be learned in an end-to-end manner together with other modules, which is optimized for the object detection task.

\begin{figure}[!tb]
	\centering
	\subfigure{
		\begin{minipage}[t]{0.3\linewidth}
			\centering
			\includegraphics[width=1.0\textwidth]{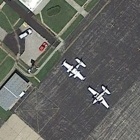}\vspace{0.04cm}
			\centering
			\includegraphics[width=1.0\textwidth]{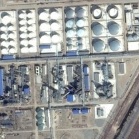}\vspace{0.04cm}
			\centering
			\includegraphics[width=1.0\textwidth]{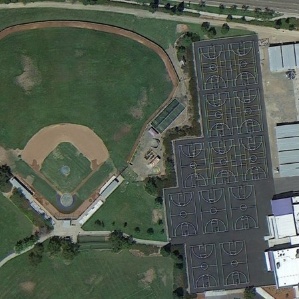}	
		\end{minipage}
		\begin{minipage}[t]{0.3\linewidth}
			\centering
			\includegraphics[width=1.0\textwidth]{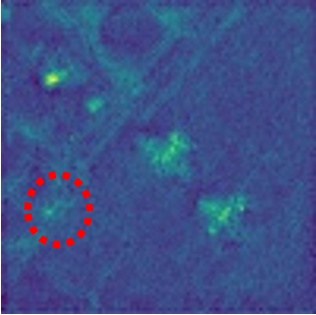}\vspace{0.04cm}
			\centering
			\includegraphics[width=1.0\textwidth]{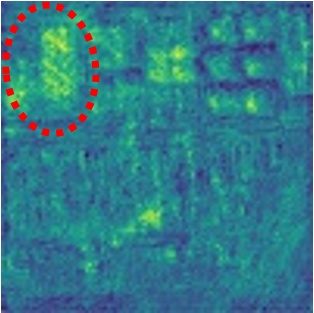}\vspace{0.04cm}
			\centering
			\includegraphics[width=1.0\textwidth]{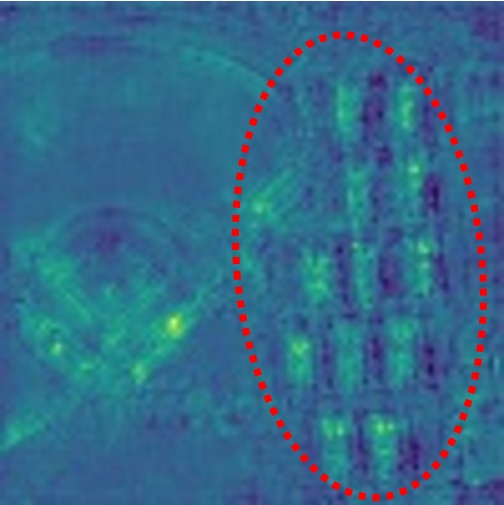}	
		\end{minipage}
		\begin{minipage}[t]{0.3\linewidth}
			\centering
			\includegraphics[width=1.0\textwidth]{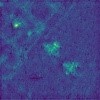}\vspace{0.04cm}
			\centering
			\includegraphics[width=1.0\textwidth]{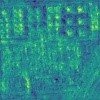}\vspace{0.04cm}
			\centering
			\includegraphics[width=1.0\textwidth]{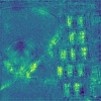}	
		\end{minipage}}
	\caption{Images (left) and their feature maps before (middle) and after (right) the instance-level denoising operation. First row: non-object with  object-like shape. Second row: inter-class feature coupling and intra-class feature boundary blurring. Third row: weak feature response.}
	\label{fig:denoise_visualize}
	%\vspace{-10pt}
\end{figure}

\subsubsection{Instance-Level Noise}\label{subsec:inld_noise}
%\textbf{Does image noise really have a huge impact on detection tasks?} Image noise often refers to unnecessary or redundant interference in images. Therefore, image noise belongs to external noise. In commonly used object detection datasets, the images tend to be relatively clean. In other words, the interference of external noise on detection is relatively small, and how to eliminate internal noise is the key.

%\textbf{What is the internal noise of the image for object detection?} We find that  have a huge impact on object detection, and we call these internal noises as \textbf{`instance noise'}. In general, we analyze aerial images, and the instance noise often occurs in the following cases on the feature map:

Instance-level noise generally refers to the mutual interference among objects, and also that from background. We discuss its properties in the following aspects. In particular, as shown in Fig. \ref{fig:denoise_visualize}, the adversary effect to object detection is especially pronounced in the feature map that calls for feature space denoising rather than on the raw input image.

1) The non-object with object-like shape has a higher response in the feature map, especially for small objects (see the top row of Fig. \ref{fig:denoise_visualize}).

2) Clutter objects that are densely arranged tend to suffer the issue for inter-class feature coupling and intra-class feature boundary blurring (see the middle row of Fig. \ref{fig:denoise_visualize}).

3) The response of object is not prominent enough surrounded by the background (see the bottom row of Fig. \ref{fig:denoise_visualize}).

\subsubsection{Mathematical Modeling of Instance-Level Denoising}
\label{sec:inld_fundamental}
To dismiss instance level noise, one can generally refer to the idea of attention mechanism, as a common way of re-weighting the convolutional response maps to highlight the important parts and suppress the uninformative ones, such as spatial attention \cite{wang2018non} and channel-wise attention \cite{hu2018squeeze}. We show that existing aerial image rotation detectors, including FADet \cite{li2019feature}, SCRDet \cite{yang2019scrdet} and CAD-Det \cite{zhang2019cad}, often use the simple attention mechanism to re-weight the output, which can be reduced into the following general form:
\begin{equation}
\small
\begin{aligned}
\mathbf{Y}  &= \mathcal{A}(\mathbf{X})\odot\mathbf{X} = \mathbf{W}_{s} \odot \mathbf{X} \odot \mathbf{W}_{c} = \mathbf{W}_{s} \odot \bigcup_{i=1}^{C}\mathbf{x}_{i} \cdot w^{i}_{c}
\label{eq:attention}
\end{aligned}
\end{equation}
where $\mathbf{X}, \mathbf{Y} \in \mathbb{R}^{C \times H \times W}$ represents two feature maps of input image. The attention function $\mathcal{A(\mathbf{X})}$ refers to the proposal output by a certain attention module e.g. \cite{wang2018non, hu2018squeeze}. Note $\odot $ is the element-wise product. $\mathbf{W}_{s} \in \mathbb{R}^{ H \times W}$ and $\mathbf{W}_{c} \in \mathbb{R}^{C}$ denote the spatial weight and channel weight. $w^{i}_{c}$ indicates the weight of the $i$-th channel, respectively. Throughout the paper, $\bigcup$ means the concatenation operation for connecting tensor along the feature map's channels.

\begin{figure}[!t]
	\centering
	\subfigure{
		\begin{minipage}[t]{0.22\linewidth}
			\centering
			\includegraphics[width=1.0\textwidth]{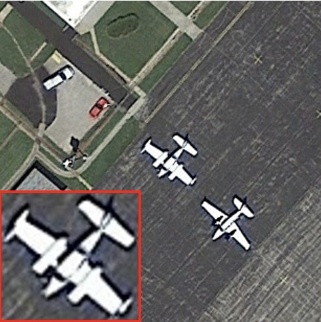}\vspace{0.04cm}
			\centering
			\includegraphics[width=1.0\textwidth]{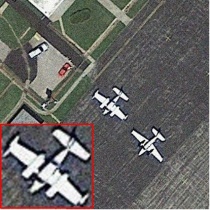}
			
	\end{minipage}
	\begin{minipage}[t]{0.22\linewidth}
			\centering
			\includegraphics[width=1.0\textwidth]{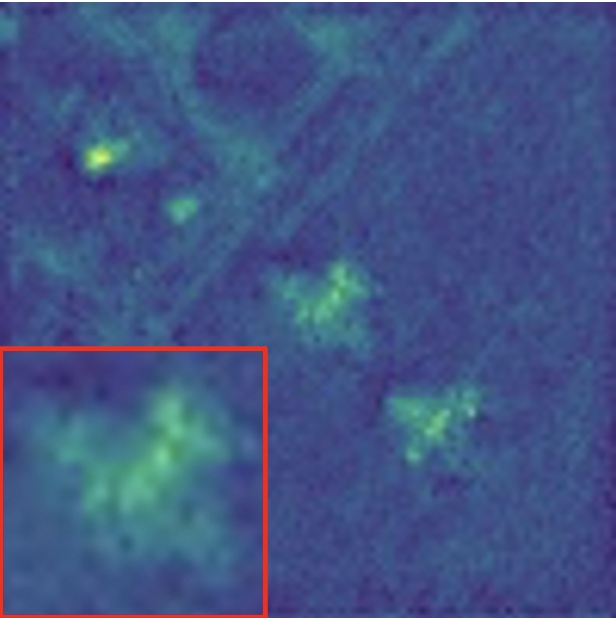}\vspace{0.04cm}
			\centering
			\includegraphics[width=1.0\textwidth]{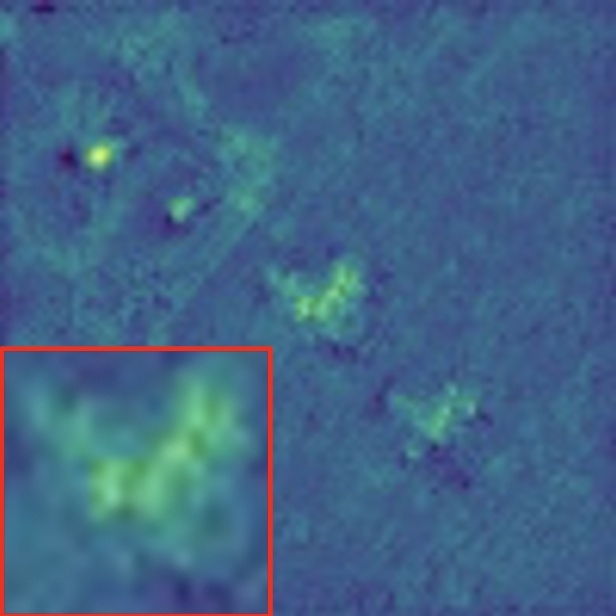}
			
	\end{minipage}
	\begin{minipage}[t]{0.22\linewidth}
			\centering
			\includegraphics[width=1.0\textwidth]{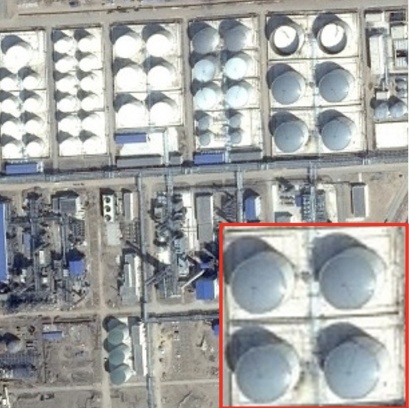}\vspace{0.04cm}
			\centering
			\includegraphics[width=1.0\textwidth]{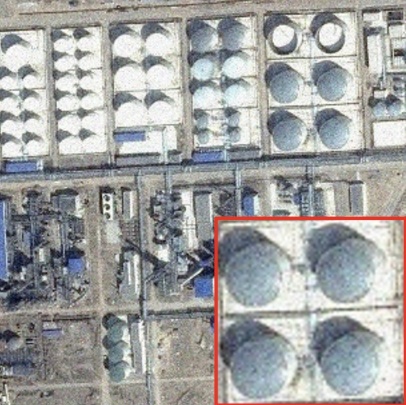}
	\end{minipage}
	\begin{minipage}[t]{0.22\linewidth}
			\centering
			\includegraphics[width=1.0\textwidth]{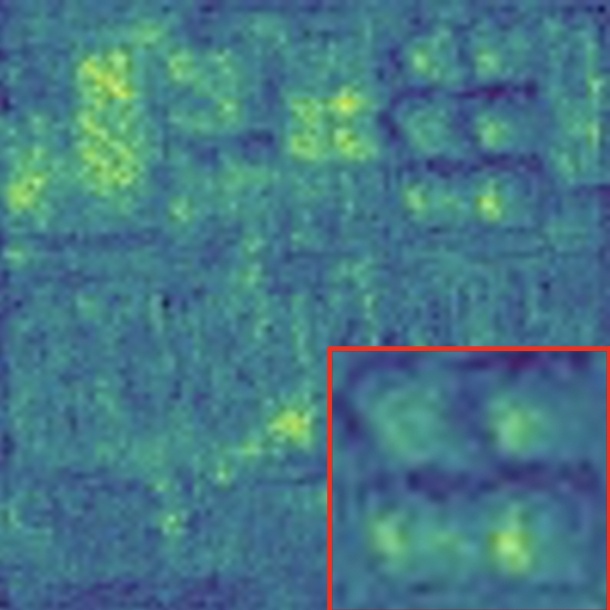}\vspace{0.04cm}
			\centering
			\includegraphics[width=1.0\textwidth]{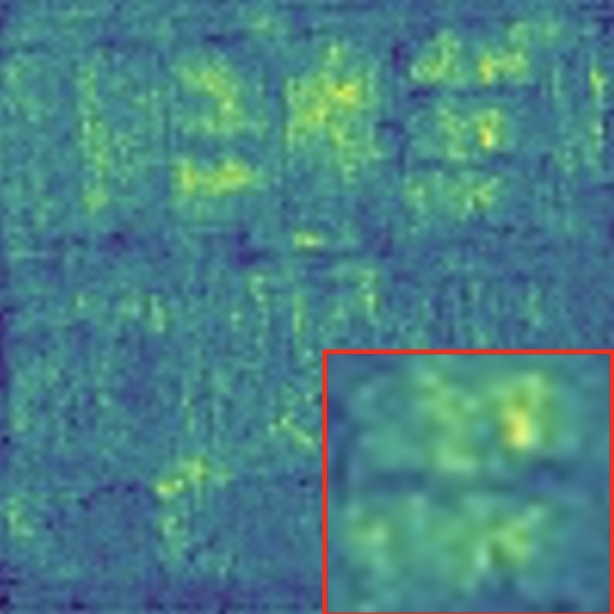}
	\end{minipage}}
	\caption{Feature maps corresponding to clean images (top) and to their noisy versions (bottom). The noise is randomly generated by a Gaussian function with a mean value of 0 and a variance of 0.005. The first and third columns: images; the rest columns: feature maps. The contrast between foreground and background in the feature map of the clean image is more obvious (second column), and the boundaries between dense objects are clearer (fourth column).}
	\label{fig:noise_visualize}
	%\vspace{-8pt}
\end{figure}

However, Eq.~\ref{eq:attention} simply distinguishes feature response between objects and background in spatial domain, and $w^{i}_{c}$ is only used to measure the importance of each channel. In other words, the interaction between intra-class objects and inter-class objects is not considered which is important for detection in complex scene. We are aimed to devise a new network that can not only distinguish object from background, but also weaken the mutual interference among objects. Specifically, we propose adding instance-level denoising (InLD) module at intermediate layers of convolutional networks. The key is to decouple the feature of different object categories into their respective channels, and meanwhile the features of objects and background are enhanced and weakened in the spatial domain, respectively.

As a result, our new formulation is as follows, which considers the total $I$ number of object categories with one additional category for background:
\begin{equation}
\small
\begin{aligned}
\mathbf{Y}  &= \mathcal{D}_{InLD}(\mathbf{X})\odot\mathbf{X} = \mathbf{W}_{InLD} \odot \mathbf{X} \\
&= \bigcup_{i=1}^{I+1}\mathbf{W}_{InLD}^{i} \odot \mathbf{X}^{i} = \bigcup_{i=1}^{I+1}\bigcup_{j=1}^{C_{i}} \mathbf{w}^{i}_{j} \odot \mathbf{x}^{i}_{j}
\label{eq:denoising}
\end{aligned}
\end{equation}
where $\mathbf{W}_{InLD} \in \mathbb{R}^{C \times H \times W}$ is a hierarchical weight. $\mathbf{W}_{InLD}^{i} \in \mathbb{R}^{C_{i} \times H \times W}$, $\mathbf{X}^{i} \in \mathbb{R}^{C_{i} \times H \times W}$ denotes the weight and feature response corresponding to the $i$-th category, and its channel number is denoted by $C_{i}$, for $C=\sum_{i=1}^{I}C_{i} + C_{bg}$. $\mathbf{w}^{i}_{j}$ and $\mathbf{x}^{i}_{j}$ denotes the weight and feature of the $i$-th category along the $j$-th channel, respectively.

As can be seen from Eq.~\ref{eq:attention} and Eq.~\ref{eq:denoising}, $\mathcal{D}_{InLD}(\mathbf{X})$ can be approximated as a combination of multiple $\mathcal{A}^{i}(\mathbf{X}^{i})$, which denotes the attention function of category $i$. Thus we have:
\begin{equation}
\small
\begin{aligned}
\mathbf{Y} & = \mathcal{D}_{InLD}(\mathbf{X})\odot\mathbf{X} = \bigcup_{i=1}^{I+1}\mathcal{A}^{i}(\mathbf{X}^{i}) \odot \mathbf{X}^{i}
\label{eq:DandA}
\end{aligned}
\end{equation}

\begin{figure}[!tb]
	\begin{center}
		\includegraphics[width=0.9\linewidth]{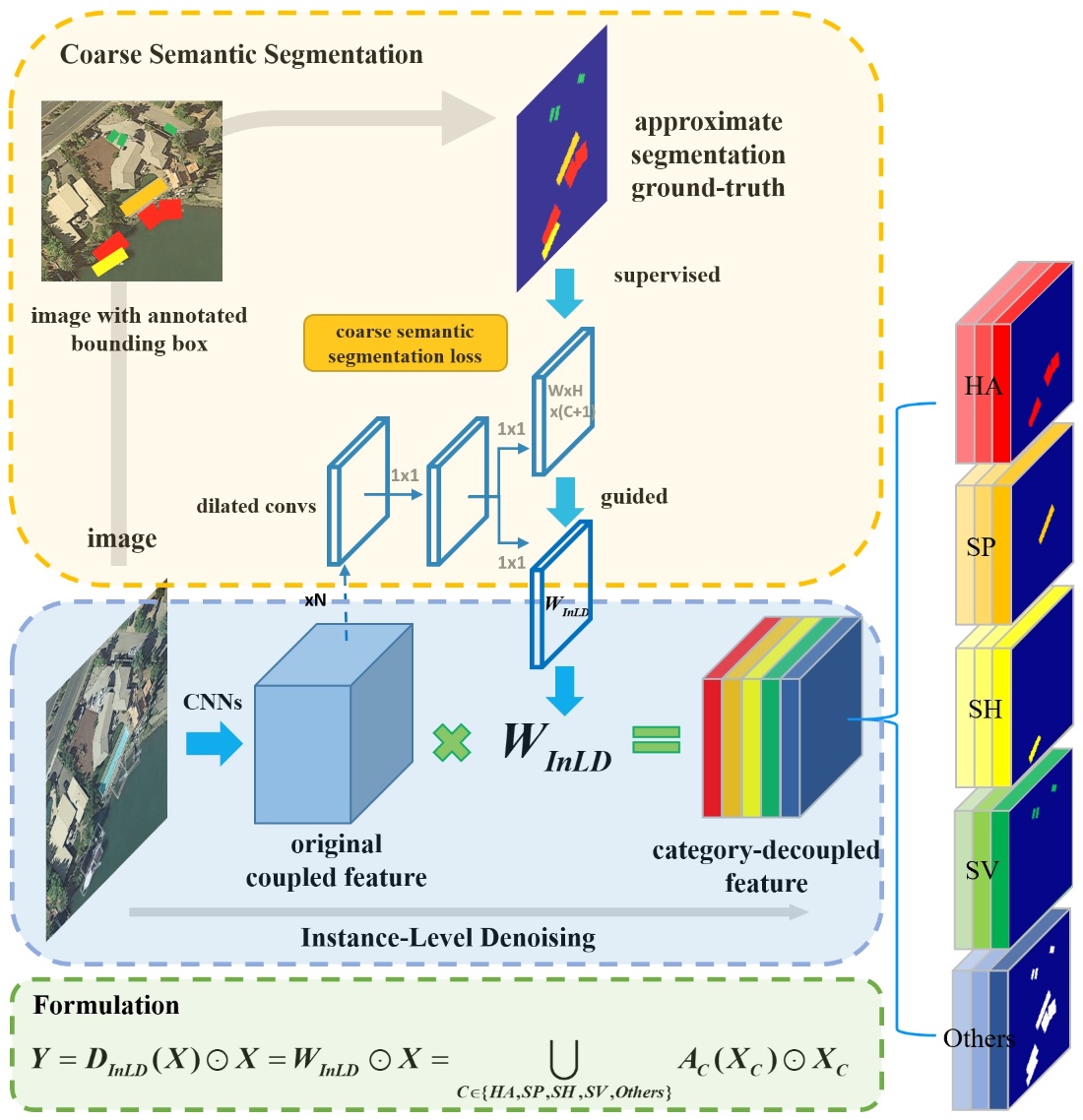}
	\end{center}
	\caption{Feature map with decoupled category-specific feature signals along channels. The abbreviation `HA', `SP', `SH', and `SV' indicate `Harbor', `Swimming pool', `Ship', and `Small vehicle', respectively. `Others' include background and unseen categories that do not appear in the image. Features of different categories are decoupled into their respective channels (top and middle), while the features of object and background are enhanced and suppressed in spatial domain, respectively (bottom).}
	\label{fig:feature_denoising}
\end{figure}

Without loss of generality, consider an image containing objects belonging to the first $I_0$ ($I_0\leq I$) categories. In this paper, we aim to decouple the above formula into three parts as concatenated to each other (see Fig. \ref{fig:feature_denoising}):
\begin{equation}
\small
\begin{aligned}
\mathbf{Y} = \underbrace{\bigcup_{i=1}^{I_0}\bigcup_{p=1}^{C_{i}} \mathbf{w}^{i}_{p} \odot \mathbf{x}^{i}_{p}}_{\text{categories in image}} \cup \underbrace{\bigcup_{j=I_0+1}^{I}\bigcup_{q=1}^{C_{j}} \mathbf{w}^{j}_{q} \odot \mathbf{x}^{j}_{q}}_{\text{categories not in image}} 
 \cup \underbrace{\bigcup_{k=1}^{C_{bg}} \mathbf{w}^{bg}_{k} \odot \mathbf{x}^{bg}_{k}}_{\text{background}}
\label{eq:denoising_}
\end{aligned}
\end{equation}

For background and unseen categories not in image, ideally the response is filtered by our devised denoising module to be as small as possible. From this perspective, Eq.~\ref{eq:denoising_} can be further interpreted by:
\begin{equation}
\small
\begin{aligned}
\mathbf{Y} &=  \underbrace{\bigcup_{i=1}^{I_0}\bigcup_{p=1}^{C_{i}} \mathbf{w}^{i}_{p} \odot \mathbf{x}^{i}_{p}}_{\text{categories in image}} \cup \underbrace{\bigcup_{j=I_0+1}^{I}\mathcal{O}_{j}}_{\text{categories not in image}} \cup \underbrace{\mathcal{O}_{bg}}_{\text{background}}
\label{eq:denoising_final}
\end{aligned}
\end{equation}
where $\mathcal{O}$ denotes tensor with small feature response one aims to achieve, for each category $\mathcal{O}_j$ and background $\mathcal{O}_{bg}$.

In the following subsection, we show how to achieve the above decoupled feature learning among categories.

\subsubsection{Implementation of Instance-Level Denoising}
\label{sec:inld_implement}
Based on the above derivations, we devise a practical neural network based implementation. Our analysis starts with the simplest case with a single channel for each category's weight $\mathbf{W}_{InLD}^{i}$ in Eq.~\ref{eq:denoising}, or namely $C_{i}=1$. In this setting, the learned weight $\mathbf{W}_{InLD}$ can be regarded as the result of semantic segmentation of the image for specific categories (a three-dimensional one-hot vector). Then more channels of weight $\mathbf{W}_{InLD}$ in $\mathcal{D}_{InLD}$ can be guided by semantic segmentation, as illustrated in Fig. \ref{fig:pipeline} and Fig. \ref{fig:feature_denoising}. In semantic segmentation task, the feature responses of each category on the previous layers of the output layer tend to be separated in the channel dimension, and the feature responses of the foreground and background in the spatial dimension are also polarized. Hence one can adopt a semantic segmentation network for the operations in Eq.~\ref{eq:denoising_final}. Another advantage for holding this semantic segmentation view is that it can be conducted in an end-to-end supervised fashion, whose learned denoising weights can be more reliable and effective than the self-attention based alternatives \cite{wang2018non,hu2018squeeze}.

%\yan{From such a segmentation perspective, we can approximate the previous layer of the output layer in the semantic segmentation task as the weight $W_{InLD}$ we need when the number of channels is not 1.} In fact, all operations in $\mathcal{D}_{InLD}$ can be implemented by  
%In simple terms, the construction of weight $\mathbf{W}_{InLD}$ is fulfilled in a way like semantic segmentation. Another 

In Fig. \ref{fig:pipeline}, we give a specific implementation as follows. The input feature map expands the receptive field by $N$ dilated convolutions \cite{yu2015multi} and a $1\times1$ convolutional layer at first. For instance, the values of $N$ take the numbers of $\{1, 1, 1, 1, 1\}$ on pyramid levels P3 to P7, respectively as set in our experiments. The feature map is then processed by two parallel $1\times1$ convolutional layers to obtain the two important outputs. One output (a three-dimensional one-hot feature map) is used to perform coarse multi-class segmentation, and the annotated bounding box in detection tasks can be used as the approximate ground truth. The hope is that this output will guide the other output into a denoising feature map.
%, with the expected function for $\mathbf{W}_{InLD}$.

As shown in Fig. \ref{fig:feature_denoising}, this denoising feature map and the original feature map are combined (by dot operation) to obtain the final decoupled feature map. The purpose is in two-folds: along the channel dimension, inter-class feature responses of different object categories (excluding the background) are basically decoupled into their respective channels; In the spatial dimension, intra-class feature boundaries are sharpened due to the feature response of the object area is enhanced and background is weakened. As such, the three issues as raised in the beginning of this subsection are alleviated.

As shown in the upper right corner of Fig. \ref{fig:pipeline}, the classification model is decomposed into two terms: objectness and category classification, as written by:
\begin{equation}
%\label{eq:class_object}
%\begin{aligned}
P(class_{i}, object) = \underbrace{P(class_{i}|object)}_{\text{category classification}} * \underbrace{P(object)}_{\text{objectness}}
\label{eq:objectness}
%\end{aligned}
\end{equation}

This probability map $P(object)$ relates to whether the anchor for each feature point is an object. While the above decoupled features are directly used for object classification $P(class_{i}|object)$ (as well as rotation regression which will be discussed in Sec.~\ref{subsec:learning}). 
%InLD weakens the feature response of non-object areas (including interferents and backgrounds), enhances that of the object area and the boundary between the objects becomes clear. Finally, we use this hierarchical feature map for classification and regression, just like the original detector does.

%For the coarse foreground binary semantic segmentation, 
%we also get a coproduct, which is a probability map.  The final class probability of the detector is:

During training, the probability map $P(object)$  will be used as a weight for the regression loss (see Eq.~\ref{eq:multitask_loss_h}), making those ambiguous positive samples get smaller weights and giving higher quality positive samples more attention. We find in the experiment that the introduction of the probability map can speed up the convergence of the model and improve the detection results, as shown in Tab. \ref{table:InLD_Ablative_Study}. 

\begin{table}[tb!]
	\centering
	\caption{Ablative study of five image level denoising settings as used in \cite{xie2019feature} on the OBB task of DOTA-v1.0 dataset.}
	\resizebox{0.48\textwidth}{!}{
		\begin{tabular}{l|c|c}
			\hline
			Base Model & Image-Level Denoising & mAP (\%)\\
			\hline
			\multirow{6}{*}{R$^3$Det \cite{ yang2021r3det}}
			& none & 65.73\\\cline{2-3}
			& bilateral, dot prod & 66.94\\
			& bilateral, gaussian & 67.03\\
			& nonlocal, dot prod & 66.82\\
			& nonlocal, gaussian & \textbf{67.68}\\
			& nonlocal, gaussian, 3x3 mean & 66.88\\
			\hline
	\end{tabular}}
	\label{table:ImLD_Ablation_Study}
\end{table}

\subsubsection{Comparison with Image-Level Denoising}\label{subsec:ImLD}
Image denoising is a fundamental task in image processing, which may impose notable impact to image recognition, as has been recently studied and verified in~\cite{xie2019feature}. Specifically, the work~\cite{xie2019feature} shows that the transformations performed by the network layers exacerbate the perturbation, and the hallucinated activations can overwhelm the activations due to true signal, which leads to worse prediction.  

Here we also study this issue in the context of aerial images through directly borrow the image level denoising model~\cite{xie2019feature}. As shown in Fig. \ref{fig:noise_visualize}, we add Gaussian noise on the raw aerial images and compare with the clean ones. The same feature map on clean and noisy images, extracted from the same channel of a res3 block in the same detection network trained on clean images are visualized. Though the noise has little effect and it is difficult to distinguish by naked eyes. However, it becomes more obvious in the feature map such that the objects are gradually submerged in the background or the boundary between the objects tends to be blurred.

Since the convolution operation and the traditional denoising filters are highly correlated, we resort to a potential solution \cite{xie2019feature} which employs convolutional layers to simulate different types of differential filters, such as non-local means, bilateral filtering, mean filtering, and median filtering. Inspired by the success of these operation in adversarial attacks \cite{xie2019feature}, in this paper we migrate and extend these differential operations for object detection. We show the generic form of ImLD in Fig. \ref{fig:pipeline}. It processes the input features by a denoising operation, such as non-local means or other variants. The denoised representation is first processed by a $1\times1$ convolutional layer, and then added to the module’s input via a residual connection. The simulation of ImLD is expressed as follows:
\begin{equation}
\small
\mathbf{Y}  = \mathcal{F}(\mathbf{X}) + \mathbf{X}
\label{eq:ImLD_EQ}
\end{equation}
where $\mathcal{F(\mathbf{X})}$ is the output by a certain filter. $\mathbf{X}$, $\mathbf{Y} \in \mathbb{R}^{C \times H \times W}$ represent the whole feature map of input image. The effect of the imposed denosing module is shown in Tab. \ref{table:ImLD_Ablation_Study}. In the following, we further show that the more notable detection improvement comes from the InLD module and its effect can well cover the image level one.

\subsection{Loss Function Design and Learning}\label{subsec:learning}
\subsubsection{Horizontal Object Detection}
%Horizontal and rotation detection settings are both considered. For rotation detection, we need to redefine the representation of the bounding box. Fig. \ref{fig:90} shows the rectangular definition of the 90 degree angle representation range \cite{yang2018automatic, yang2019scrdet, yang2021r3det,qian2021learning,yang2018position}. $\theta$ denotes the acute angle to the x-axis, and for the other side we refer it as $w$. Note this definition is also officially adopted by OpenCV \url{https://opencv.org/}.

For horizontal detection, regressing the bounding box by:
\begin{equation}
\small
\begin{aligned}
t_{x}&=(x-x_{a})/w_{a}, t_{y}=(y-y_{a})/h_{a} \\
t_{w}&=\log(w/w_{a}), t_{h}=\log(h/h_{a}), \\ %t_{\theta}&=\theta-\theta_{a} \quad (only\;for\;rotation\;detection)
t_{x}^{'}&=(x_{}^{'}-x_{a})/w_{a}, t_{y}^{'}=(y_{}^{'}-y_{a})/h_{a} \\
t_{w}^{'}&=\log(w_{}^{'}/w_{a}), t_{h}^{'}=\log(h_{}^{'}/h_{a}) % t_{\theta}^{'}&=\theta_{}^{'}-\theta_{a} \quad (only\;for\;rotation\;detection)
\label{eq:regression}
\end{aligned}
\end{equation}
where $x, y, w, h$ denote the box's center coordinates, width, and height, respectively. Variables $x, x_{a}, x^{'}$ are for the ground-truth box, anchor box, and predicted box, respectively (likewise for $y,w,h$).

The multi-task loss of horizontal detection is defined as:
\begin{equation}
\small
\begin{aligned}
L_{h} = &\frac{\lambda_{reg}}{N}\sum_{n=1}^{N}t_{n}^{'} \cdot p(object_{n})\sum_{j\in\{x,y,w,h\}}L_{reg}(v_{nj}^{'},v_{nj}) \\
& + \frac{\lambda_{cls}}{N}\sum_{n=1}^{N}L_{cls}(p_{n},t_{n}) + \frac{\lambda_{InLD}}{h\times w}\sum_{i}^{h}\sum_{j}^{w}L_{InLD}(u_{ij}^{'},u_{ij})
\label{eq:multitask_loss_h}
\end{aligned}
\end{equation}
where $N$ indicates the number of anchors, $t_{n}^{'}$ is a binary value ($t_{n}^{'}=1$ for foreground and $t_{n}^{'}=0$ for background, no regression for background). $p(object_{n})$ indicates the probability that the current anchor is the object. $v_{nj}^{'}$ denotes the predicted offset vectors of the n-th anchor, $v_{nj}$ is the targets vector between n-th anchor and ground-truth it matches. $t_{n}$ represents the label of object, $p_{n}$ is the probability distribution of various classes calculated by sigmoid function. $u_{ij}$, $u_{ij}^{'}$ denote the label and predict of mask's pixel respectively. The hyper-parameter $\lambda_{reg}$, $\lambda_{cls}$, $\lambda_{InLD}$ control the trade-off and are set to 1 by default. The classification loss $L_{cls}$ is focal loss \cite{lin2017focal}. The regression loss $L_{reg}$ is smooth L1 loss as defined in \cite{girshick2015fast}, and the InLD loss $L_{InLD}$ is pixel-wise softmax cross-entropy.

\begin{figure}[!tb]
	\centering
	\subfigure{
		\begin{minipage}[t]{0.9\linewidth}
			\centering
			\includegraphics[width=1.0\textwidth]{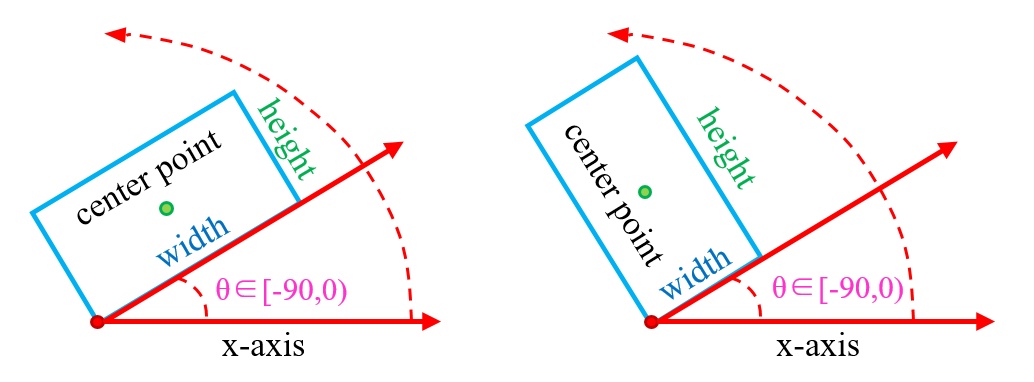}
			\centering
			\label{fig:90}
	\end{minipage}} \\
	%\vspace{-15pt}
% 	\subfigure[Long side definition]{
% 		\begin{minipage}[t]{0.9\linewidth}
% 			\centering
% 			\includegraphics[width=1.0\textwidth]{figure_mini/180_.jpg}
% 			\centering
% 			\label{fig:180}
% 	\end{minipage}}
	\caption{Rotation box definitions (OpenCV definition). $\theta $ denotes the acute angle to the x-axis, and for the other side we refer it as $w$. The range of angle representation is $[-90,0)$.}
	\label{fig:definition}
	%\vspace{-10pt}
\end{figure}
\begin{figure}[!tb]
	\centering
	\subfigure[Ideal case.]{
		\begin{minipage}[t]{0.58\linewidth}
			\centering
			\includegraphics[width=1.0\textwidth]{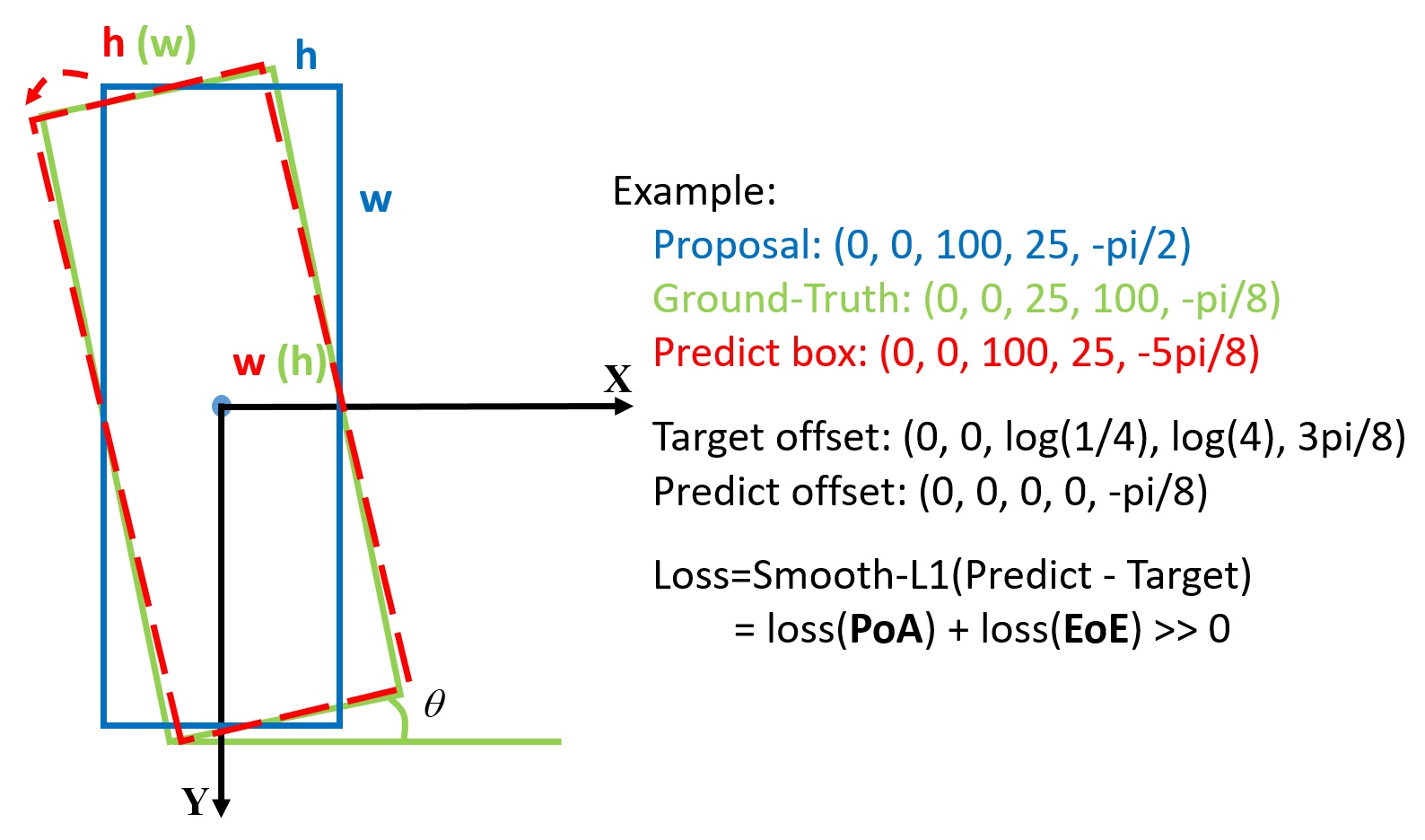}
			\centering
			\label{fig:example1}
	\end{minipage}}
	\subfigure[Actual case.]{
		\begin{minipage}[t]{0.35\linewidth}
			\centering
			\includegraphics[width=1.0\textwidth]{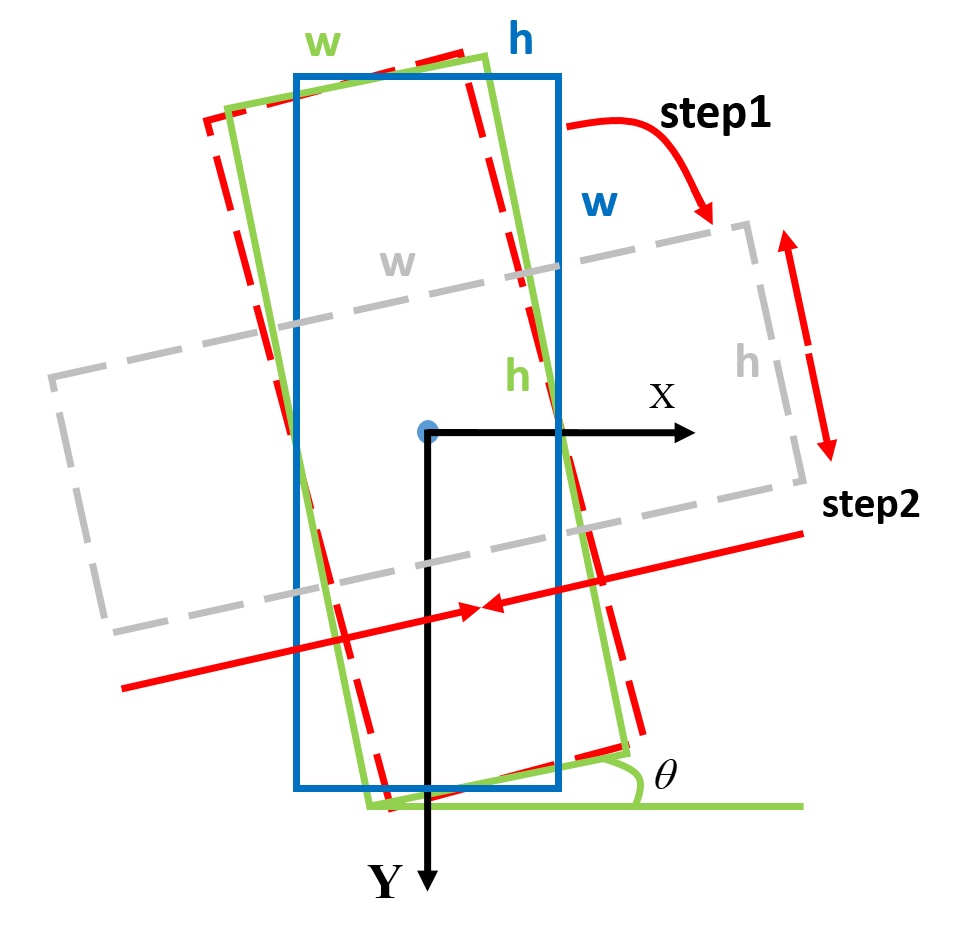}
			\centering
			\label{fig:example2}
	\end{minipage}}
	\caption{Boundary discontinuity of angle regression. Blue, green, red bounding box denotes anchor/proposal, ground-truth, prediction box.}
	\label{fig:example}
\end{figure}
\begin{figure}[!tb]
	\centering
	\subfigure[Smooth L1 loss.]{
		\begin{minipage}[t]{0.4\linewidth}
			\centering
			\includegraphics[width=1.0\textwidth]{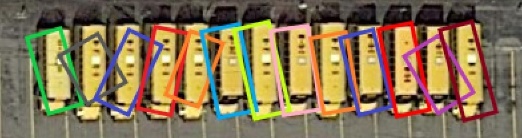}
			\centering
			\label{fig:bad_case}
	\end{minipage}}
	\subfigure[IoU-smooth L1 loss.]{
		\begin{minipage}[t]{0.4\linewidth}
			\centering
			\includegraphics[width=1.0\textwidth]{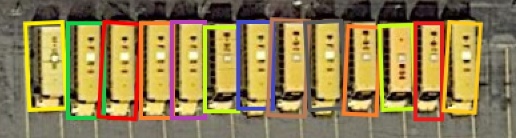}
			\centering
			\label{fig:good_case}
	\end{minipage}}
	\caption{Detection results by two losses. For this dense arrangement case, the angle estimation error will also make the classification even harder.}
	\label{fig:cases}
\end{figure}
\begin{table}[tb!]
	\centering
	\caption{Ablative study for speed and accuracy of InLD on OBB task of DOTA. Binary-Mask and Multi-Mask refer to binary and multi-class semantic segmentation, respectively. Coproduct denotes multiplying the objectness term $P(object)$ or not in Eq.~\ref{eq:objectness}.}
	\resizebox{0.48\textwidth}{!}{
		\begin{tabular}{l|c|c|c|c}
			\hline
			Base Model & Mask Type & Coproduct & FPS & mAP (\%)\\
			\hline
			\multirow{4}{*}{R$^3$Det \cite{ yang2021r3det}}
			& null & $\times$ & 14 & 65.73\\
			& Binary-Mask & $\times$ & 13.5 & 68.12\\
			& Multi-Mask & $\times$ & 13 & 69.43\\
			& Multi-Mask & $\surd$ & 13 & \textbf{69.81}\\
			\hline
	\end{tabular}}
	\label{table:InLD_Ablative_Study}
\end{table}
\subsubsection{Rotation Object Detection}
\label{sec:rod}
In contrast, we need to redefine the representation of the bounding box. Fig. \ref{fig:90} shows the rectangular definition of the 90 degree angle representation range \cite{yang2018automatic, yang2019scrdet, yang2021r3det,qian2021learning,yang2018position}. $\theta$ denotes the acute angle to the x-axis, and for the other side we refer it as $w$. Note this definition is also officially adopted by OpenCV\footnote{\url{https://opencv.org/}}. 

Rotation detection needs to carefully address the boundary problem. In particular, there exists the boundary problem for the angle regression, as shown in Fig. \ref{fig:example1}. It shows that an ideal form of regression (the blue box rotates counterclockwise to the red box), but the loss of this situation is very large due to the periodicity of angular (PoA) and exchangeability of edges (EoE). Therefore, the model has to be regressed in other complex forms like in Fig. \ref{fig:example2} (such as the blue box rotating clockwise while scaling $w$ and $h$), increasing the difficulty of regression, as shown in Fig. \ref{fig:bad_case}.

As for the regression equation of $\theta$, we use two forms as the baseline to be compared:
\begin{itemize}
	\item Direct regression (default in this paper), namely Reg. ($\Delta \theta$). The model directly predicts the angle offset $t_{\theta}^{*}$:
	\begin{equation}
    \begin{aligned}
    	t_{\theta}=&(\theta-\theta_{a})\cdot\pi/180 \\ 
    	t_{\theta}^{'}=&(\theta_{}^{'}-\theta_{a})\cdot\pi/180
    \end{aligned}
    \end{equation}
	\item Indirect regression, marked as Reg.$^*$ ($\sin{\theta}$, $\cos{\theta}$). It predicts two vectors ($t_{\sin\theta}^{*}$ and $t_{\cos\theta}^{*}$) to match the two targets from ground truth ($t_{\sin\theta}$ and $t_{\cos\theta}$):
	\begin{equation}
    \begin{aligned}
        t_{\sin\theta} =& \sin{(\theta\cdot\pi/180)}, t_{\cos\theta} = \cos{(\theta\cdot\pi/180)} \\ t_{\sin\theta}^{'}=&\sin{(\theta^{'}\cdot\pi/180)}, t_{\cos\theta}^{'}=\cos{(\theta^{'}\cdot\pi/180)}
    \end{aligned}
    \end{equation}
\end{itemize}
To ensure that $t_{\sin\theta}^{'2}+t_{\cos\theta}^{'2}=1$ is satisfied, we perform the following normalization processing:
\begin{equation}
    \small
    \begin{aligned}
    t_{\sin\theta}^{'}=&\frac{t_{\sin\theta}^{'}}{\sqrt{t_{\sin\theta}^{'2}+t_{\cos\theta}^{'2}}} \\ t_{\cos\theta}^{'}=&\frac{t_{\cos\theta}^{'}}{\sqrt{t_{\sin\theta}^{'2}+t_{\cos\theta}^{'2}}}
    \end{aligned}
\end{equation}

It should be noted that indirect regression is a simpler way to avoid boundary problems.

In order to better solve the boundary problem, we introduce the IoU constant factor $\frac{|-\log(IoU)|}{|L_{reg}(v_{j}^{'},v_{j})|}$ in the traditional smooth L1 loss, as shown in Eq. \ref{eq:multitask_loss_r}. This new loss function is named IoU-smooth L1 loss. It can be seen that in the boundary case, the loss function is approximately equal to $|-\log(IoU)|\approx0$, eliminating the sudden increase in loss caused by $|L_{reg}(v_{j}^{'},v_{j})|$, as shown in Fig. \ref{fig:good_case}. The new regression loss can be divided into two parts: $\frac{L_{reg}(v_{j}^{'},v_{j})}{|L_{reg}(v_{j}^{'},v_{j})|}$ determines the direction of gradient propagation, and $|-\log(IoU)|$ for the magnitude of  gradient. In addition, using IoU to optimize location accuracy is consistent with IoU-dominated metric, which is more straightforward and effective than coordinate regression.
\begin{equation}
\small
\begin{aligned}
L_{r}  =& \frac{\lambda_{reg}}{N}\sum_{n=1}^{N}t_{n}^{'} \cdot p(object_{n}) \\
 & \cdot \sum_{j\in\{x,y,w,h,\theta\}}\frac{L_{reg}(v_{nj}^{'},v_{nj})}{|L_{reg}(v_{nj}^{'},v_{nj})|}\left|-\log(IoU)\right| \\	
& + \frac{\lambda_{cls}}{N}\sum_{n=1}^{N}L_{cls}(p_{n},t_{n})  + \frac{\lambda_{InLD}}{h\times w}\sum_{i}^{h}\sum_{j}^{w}L_{InLD}(u_{ij}^{'},u_{ij})
\label{eq:multitask_loss_r}
\end{aligned}
\end{equation}
where IoU is the overlap of prediction and ground-truth.

\begin{table}[tb!]
	\centering
	\caption{Ablative study by accuracy (\%) of the number of dilated convolution on pyramid levels and the InLD loss $L_{InLD}$ in InLD on OBB task of DOTA. It can be found that supervised learning is the main contribution of InLD rather than more convolution layers.}
	\resizebox{0.45\textwidth}{!}{
	    \begin{threeparttable}
		\begin{tabular}{cccc}
			\hline
			 \multicolumn{2}{c}{InLD} & \multirow{2}{*}{RetinaNet-H \cite{ yang2021r3det}}& \multirow{2}{*}{R$^3$Det \cite{ yang2021r3det}}\\
			\cline{1-2}
			dilated convolution \cite{yu2015multi} & $L_{InLD}$ & &\\
			\hline
			-- & -- & 62.21 & 65.73 \\
			\{4,4,3,2,2\} & $\times$ & 62.36 & 66.62\\
			\{1,1,1,1,1\} & $\surd$ & 65.40 & \textbf{69.81}\\
			\{4,4,3,2,2\} & $\surd$ & \textbf{65.52} & 69.07\\
			\hline
	\end{tabular}
	\end{threeparttable}}
	\label{table:InLD}
\end{table}

\begin{table}[tb!]
	\centering
	\caption{Detailed ablative study by accuracy (\%) of the effect of InLD on two traffic light datasets. Note the category `wait on' is only available in our collected S$^2$TLD dataset as released by this paper.}
	\resizebox{0.48\textwidth}{!}{
		\begin{tabular}{l|c|c|c|c|c|c|c|c}
			\hline
			Dataset & Base Model & InLD & red & yellow & green & off & wait on & mAP \\
			\hline
			\multirow{4}{*}{S$^2$TLD} & \multirow{2}{*}{RetinaNet \cite{lin2017focal}} & $\times$ & 97.94 & 88.63 & 97.17 & 90.13 & 92.40 & 93.25 \\
			& & $\surd$ & 98.15 & 87.66 & 97.12 & 93.88 & 93.75 & \textbf{94.11} \\
			\cline{2-9}
			& \multirow{2}{*}{FPN \cite{lin2017feature}} & $\times$ & 97.98 & 87.55 & 97.42 & 93.42 & 98.31 & 94.93 \\
			& & $\surd$ & 98.04 & 92.84 & 97.69 & 92.06 & 99.08 & \textbf{95.94}\\
			\hline
			\multirow{4}{*}{BSTLD \cite{behrendt2017deep}} & \multirow{2}{*}{RetinaNet \cite{lin2017focal}} & $\times$ & 69.91 & 19.71 & 77.11 & 22.33 & -- & 47.26 \\
			& & $\surd$ & 70.50 & 24.05 & 77.16 & 22.51 & -- & \textbf{48.56}\\
			\cline{2-9}
			& \multirow{2}{*}{FPN \cite{lin2017feature}} & $\times$ & 89.27 & 47.82 & 92.01 & 40.73 & -- & 67.46 \\
			& & $\surd$ & 89.88 & 49.93 & 92.42 & 42.45 & -- & \textbf{68.67} \\
			\hline
	\end{tabular}}
	\label{table:STLD}
\end{table}

\begin{figure*}[!tb]
	\centering
	\subfigure[red, green, off]{
		\begin{minipage}[t]{0.19\linewidth}
			\centering
			\includegraphics[width=1.0\textwidth]{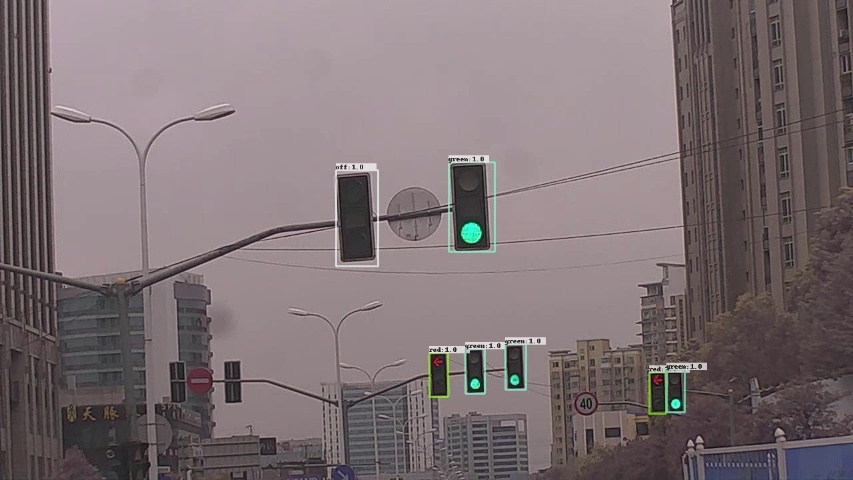}
			\centering
	\end{minipage}}
	\subfigure[red, green, wait on]{
		\begin{minipage}[t]{0.19\linewidth}
			\centering
			\includegraphics[width=1.0\textwidth]{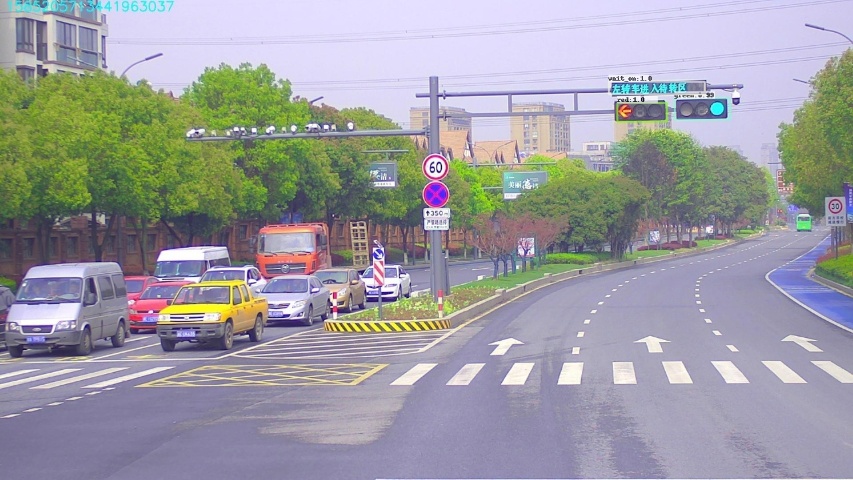}
	\end{minipage}}
	\subfigure[red]{
		\begin{minipage}[t]{0.19\linewidth}
			\centering
			\includegraphics[width=1.0\textwidth]{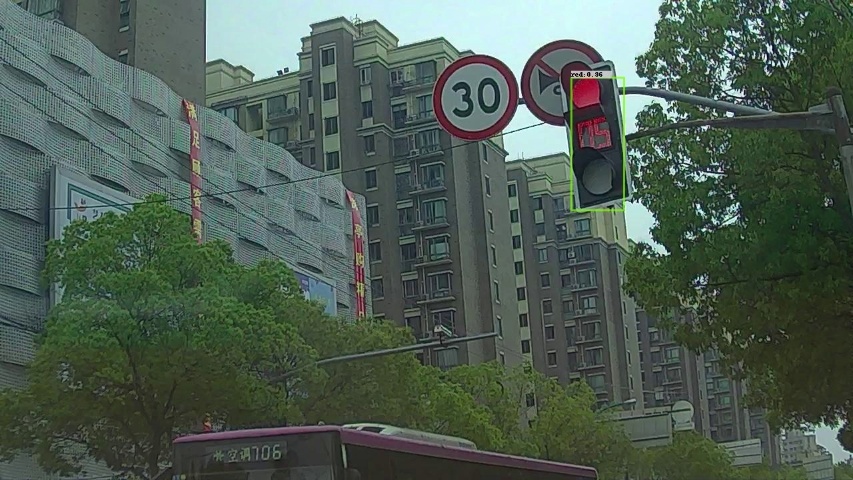}	
	\end{minipage}}
	\subfigure[red, green, off]{
		\begin{minipage}[t]{0.19\linewidth}
			\centering
			\includegraphics[width=1.0\textwidth]{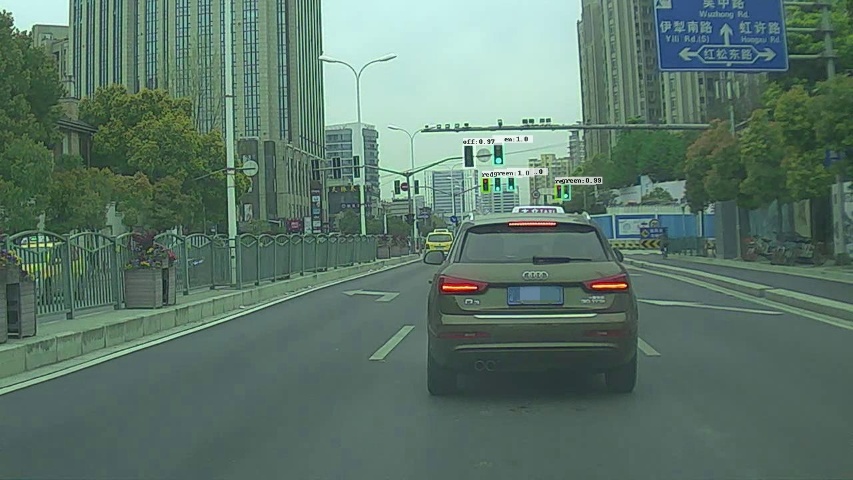}
	\end{minipage}}
	\subfigure[red, yellow]{
		\begin{minipage}[t]{0.19\linewidth}
			\centering
			\includegraphics[width=1.0\textwidth]{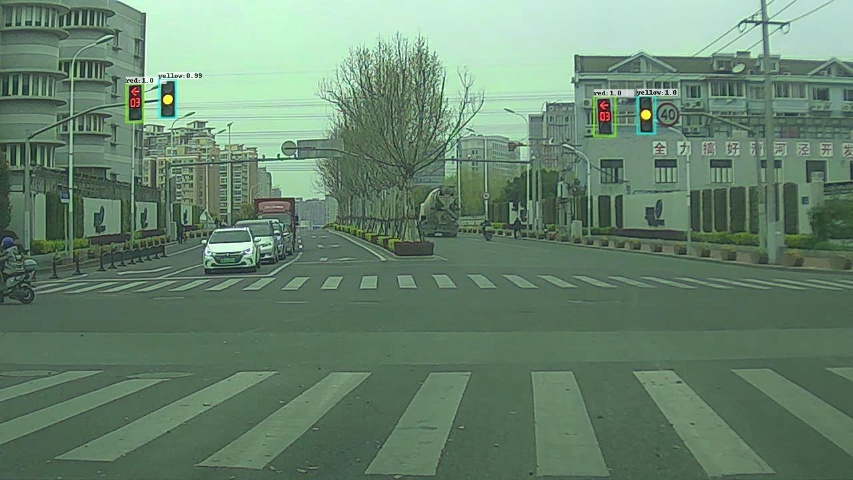}	
	\end{minipage}}\\
	\subfigure[red]{
		\begin{minipage}[t]{0.19\linewidth}
			\centering
			\includegraphics[width=1.0\textwidth]{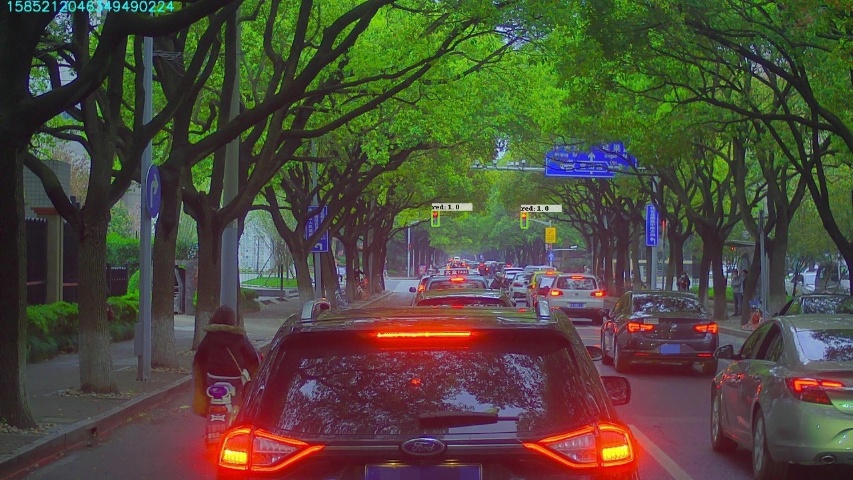}
			\centering
	\end{minipage}}
	\subfigure[red, green, wait on]{
		\begin{minipage}[t]{0.19\linewidth}
			\centering
			\includegraphics[width=1.0\textwidth]{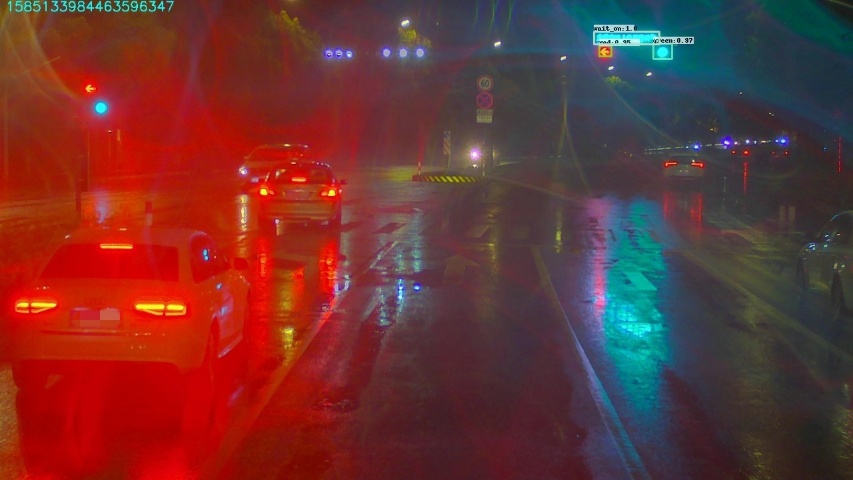}
	\end{minipage}}
	\subfigure[red]{
		\begin{minipage}[t]{0.19\linewidth}
			\centering
			\includegraphics[width=1.0\textwidth]{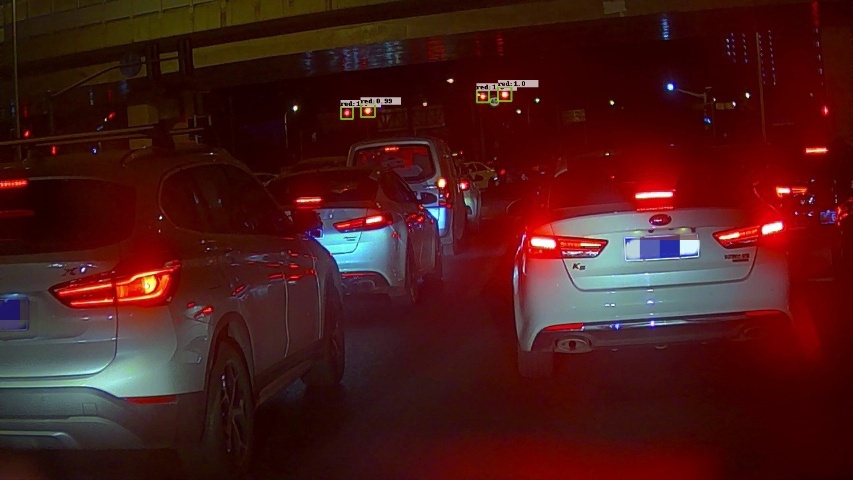}	
	\end{minipage}}
	\subfigure[red]{
		\begin{minipage}[t]{0.19\linewidth}
			\centering
			\includegraphics[width=1.0\textwidth]{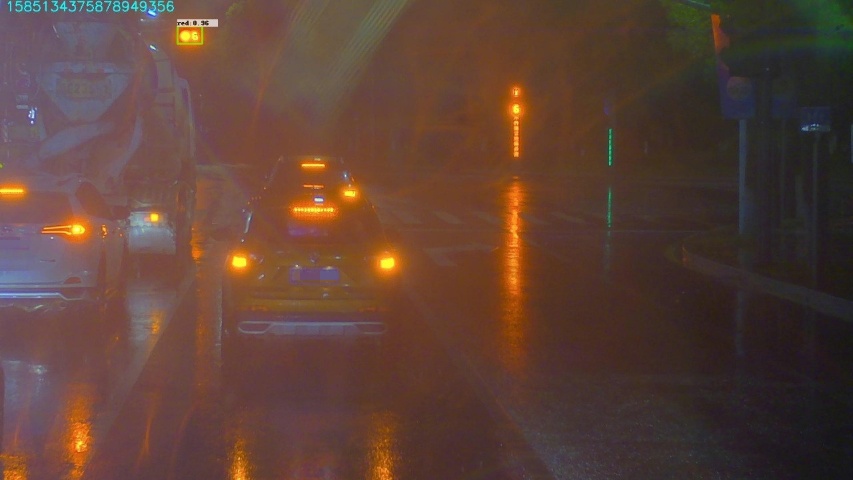}
	\end{minipage}}
	\subfigure[green]{
		\begin{minipage}[t]{0.19\linewidth}
			\centering
			\includegraphics[width=1.0\textwidth]{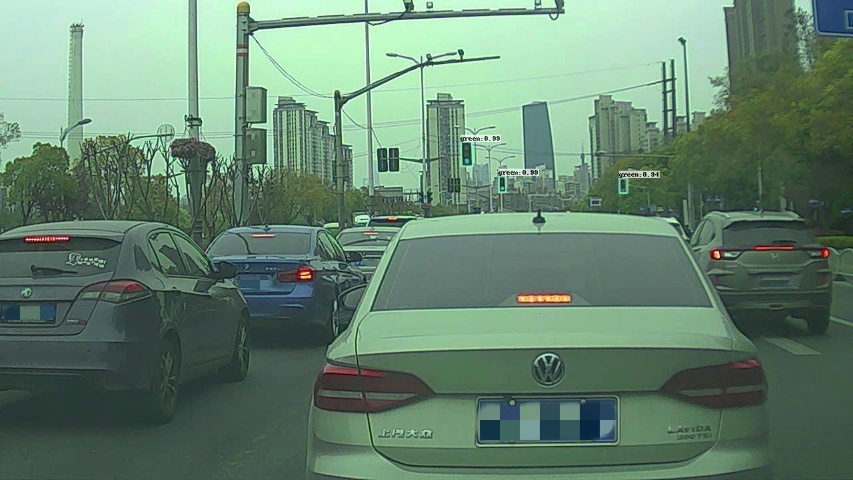}	
	\end{minipage}}
	\caption{Illustrations of the five categories and different lighting and weather conditions in our collected S$^2$TLD dataset as released in the paper.}
	\label{fig:S2TLD_VIS}
\end{figure*}

\begin{table}[tb!]
	\centering
	\caption{Ablative study by accuracy (\%) of ImLD, InLD and their combination (numbers in bracket denote relative improvement against using InLD alone) on different datasets and different detection tasks.}
	\resizebox{0.48\textwidth}{!}{
		\begin{tabular}{l|c|c|c|c|c}
			\hline
			Dataset and task & Base Model & Baseline &  ImLD & InLD & ImLD + InLD \\
			\hline
			\multirow{3}{*}{DOTA-v1.0 OBB\cite{xia2018dota}}
			& RetinaNet-H \cite{ yang2021r3det} & 62.21 & 62.39 & 65.40 & \textbf{65.62 (+0.22)} \\
			& RetinaNet-R \cite{ yang2021r3det} & 61.94 & 63.96 & 64.52 & \textbf{64.60 (+0.08)} \\
			& R$^3$Det \cite{ yang2021r3det} & 65.73 & 67.68 & 69.81 & \textbf{69.95 (+0.14)}\\
			\hline
			\multirow{1}{*}{DOTA-v1.0 HBB \cite{xia2018dota}}
			& RetinaNet \cite{lin2017focal} & 67.76 & 68.05 & 68.33 & \textbf{68.50 (+0.17)} \\
			\hline
			\multirow{2}{*}{DIOR \cite{li2020object}}
			& RetinaNet \cite{lin2017focal} & 68.05 & 68.42 & \textbf{69.36} & 69.35 (-0.01)\\
			& FPN \cite{lin2017feature} & 71.74 & 71.83 & 73.21 & \textbf{73.25 (+0.04)}\\
			\hline
			\multirow{1}{*}{ICDAR2015 \cite{karatzas2015icdar}}
			& RetinaNet-H \cite{ yang2021r3det} & 77.13 & -- & \textbf{78.68} & --\\
			\hline
			\multirow{2}{*}{COCO \cite{lin2014microsoft}}
			& FPN \cite{lin2017feature} & 36.1 & -- & \textbf{37.2} & --\\
			& RetinaNet \cite{lin2017focal} & 34.4 & -- & \textbf{35.8} & --\\
			\hline
			\multirow{1}{*}{S$^2$TLD}
			& RetinaNet \cite{lin2017focal} & 93.25 & -- & \textbf{94.11} & --\\
			\hline
	\end{tabular}}
	\label{table:ImLD_and_InLD}
\end{table}

\section{Experiments}	
\label{sec:experiment}
Experiments are performed on a server with GeForce RTX 2080 Ti and 11G memory. We first give the description of the dataset, and then use these datasets to verify the advantage of the proposed method. Source code is available at \url{https://github.com/SJTU-Thinklab-Det/DOTA-DOAI}.

\begin{table*}[tb!]
	\centering
	\caption{Ablative study by accuracy (\%) of each component in our method on the OBB task of DOTA-v1.0 dataset. For RetinaNet, `H' and `R' represents the horizontal and rotation anchors, respectively.}
	\resizebox{1.0\textwidth}{!}{
		\begin{tabular}{l|c|c|c|c|c|c|c|c|c|c|c|c|c|c|c|c|c|c|c}
			\hline
			Base Method & Backbone & InLD & Data Aug. &  PL &  BD &  BR &  GTF &  SV &  LV &  SH &  TC &  BC &  ST &  SBF &  RA &  HA &  SP &  HC &  mAP\\
			\hline
			\multirow{2}{*}{RetinaNet-H \cite{ yang2021r3det}}
			& ResNet50 & $\times$ & $\times$ & 88.87 & 74.46 & 40.11 & 58.03 & 63.10 & 50.61 & 63.63 & \textbf{90.89} & 77.91 & 76.38 & 48.26 & 55.85 & 50.67 & 60.23 & 34.23 & 62.22 \\
			& ResNet50 & $\surd$ & $\times$ & 88.83 & 74.70 & 40.80 & 65.85 & 59.76 & 53.51 & 67.38 & 90.82 & 78.49 & 80.52 & 52.02 & 59.77 & 53.56 & 66.80 & 48.24 & 65.40 \\
			\hline
			\multirow{2}{*}{RetinaNet-R \cite{ yang2021r3det}}
			& ResNet50 & $\times$ & $\times$ & 88.92 & 67.67 & 33.55 & 56.83 & 66.11 & 73.28 & 75.24 & 90.87 & 73.95 & 75.07 & 43.77 & 56.72 & 51.05 & 55.86 & 21.46 & 62.02 \\
			& ResNet50 & $\surd$ & $\times$ & 88.96 & 70.77 & 33.30 & 62.02 & 66.35 & 75.69 & 73.49 & 90.84 & 78.73 & 77.21 & 47.54 & 55.59 & 51.52 & 58.06 & 37.65 & 64.52 \\
			\hline
			\multirow{5}{*}{R$^3$Det \cite{ yang2021r3det}}
			& ResNet50 & $\times$ & $\times$ & 88.78 & 74.69 & 41.94 & 59.88 & 68.90 & 69.77 & 69.82 & 90.81 & 77.71 & 80.40 & 50.98 & 58.34 & 52.10 & 58.30 & 43.52 & 65.73 \\	
			& ResNet152 & $\times$ & $\surd$ & 89.24 & 80.81 & \textbf{51.11} & 65.62 & 70.67 & 76.03 & 78.32 & 90.83 & 84.89 & 84.42 & 65.10 & 57.18 & 68.10 & 68.98 & 60.88 & 72.81\\
			% & ResNet50 & $\surd$ & $\times$ & $\times$ & 88.92 & 76.63 & 44.74 & 61.92 & 69.35 & 73.62 & 74.84 & 90.85 & 79.87 & 81.30 & 47.96 & 59.65 & 55.67 & 60.25 & 49.59 & 67.68 \\
			& ResNet50 & $\surd$ & $\times$ & 88.63 & 75.98 & 45.88 & 65.45 & 69.74 & 74.09 & 78.30 & 90.78 & 78.96 & 81.28 & 56.28 & 63.01 & 57.40 & 68.45 & 52.93 & 69.81 \\
			%& ResNet50 & $\surd$ & $\surd$ & $\times$ & 88.91 & 76.65 & 45.44 & 62.29 & 69.90 & 75.37 & 76.21 & 90.83 & 80.89 & 81.90 & 53.85 & 58.94 & 55.76 & 64.06 & 53.32 & 68.95 \\
			%& ResNet101 & $\times$ & $\surd$ & $\times$ & 89.26 & 82.12 & 50.30 & 63.97 & 71.58 & 76.67 & 79.35 & 90.64 & 78.26 & 84.59 & 64.46 & 63.76 & 66.81 & 60.84 & 50.56 & 71.55 \\
			& ResNet101 & $\surd$ & $\surd$ & \textbf{89.25} & 83.30 & 49.94 & 66.20 & \textbf{71.82} & 77.12 & \textbf{79.53} & 90.65 & 82.14 & \textbf{84.57} & 65.33 & \textbf{63.89} & 67.56 & 68.48 & 54.89 & 72.98 \\
			& ResNet152 & $\surd$ & $\surd$ & 89.20 & \bf{83.36} & 50.92 & \textbf{68.17} & 71.61 & {\bf80.23} & 78.53 & 90.83 & \textbf{86.09} & 84.04 & {\bf65.93} & 60.80 & {\bf68.83} & {\bf71.31} & {\bf66.24} & {\bf74.41} \\
			\hline
	\end{tabular}}
	\label{table:dota_ablation_study}
\end{table*}

\begin{table*}[!tb]
	\centering
	\caption{Ablative study by accuracy (\%) of IoU-Smooth L1 loss by using it or not in the three methods on the OBB task of DOTA-v1.0 dataset. Numbers in bracket denote the relative improvement by using the proposed IoU-Smooth L1 loss.}
	\resizebox{0.98\textwidth}{!}{
		\begin{tabular}{l|c|c|c|c|c|c|c|c|c|c|c|c|c|c|c|c|c|c|c}
			\hline
			Method & Backbone & IoU-Smooth L1 & InLD & PL &  BD &  BR &  GTF &  SV &  LV &  SH &  TC &  BC &  ST &  SBF &  RA &  HA &  SP &  HC &  mAP \\
			\hline
			\multirow{2}{*}{RetinaNet-R \cite{ yang2021r3det}}
			& ResNet50 & $\times$ & $\times$ & 88.92 & 67.67 & 33.55 & 56.83 & 66.11 & 73.28 & 75.24 & 90.87 & 73.95 & 75.07 & 43.77 & 56.72 & 51.05 & 55.86 & 21.46 & 62.02 \\
			& ResNet50 & $\surd$ & $\times$ & 89.27 & 74.93 & 37.01 & 64.49 & 66.00 & 75.87 & 77.75 & 90.76 & 80.35 & 80.31 & 54.75 & 61.17 & 61.07 & 64.78 & 51.24 & 68.65 \textbf{(+6.63)} \\
			\hline
			\multirow{2}{*}{SCRDet \cite{yang2019scrdet}}
			& ResNet101 & $\times$ & $\times$ & 89.65 & 79.51 & 43.86 & 67.69 & 67.41 & 55.93 & 64.86 & 90.71 & 77.77 & 84.42 & 57.67 & 61.38 & 64.29 & 66.12 & 62.04 & 68.89 \\
			& ResNet101 & $\surd$ & $\times$ & 89.41 & 78.83 & 50.02 & 65.59 & 69.96 & 57.63 & 72.26 & 90.73 & 81.41 & 84.39 & 52.76 & 63.62 & 62.01 & 67.62 & 61.16 & 69.83 \textbf{(+0.94)} \\	
			\hline
			\multirow{2}{*}{FPN \cite{lin2017feature}}
			& ResNet101 &$\times$ & $\surd$ & 90.25 & 85.24 & 55.18 & 73.24 & 70.38 & 73.77 & 77.00 & 90.77 & 87.74 & 86.63 & 68.89 & 63.45 & 72.73 & 67.96 & 60.23 & 74.90 \\
			& ResNet101 & $\surd$ & $\surd$  & 89.77 & 83.90 & 56.30 & 73.98 & 72.60 & 75.63 & 82.82 & 90.76 & 87.89 & 86.14 & 65.24 & 63.17 & 76.05 & 68.06 & 70.24 & 76.20 \textbf{(+1.30)}  \\
			\hline
	\end{tabular}}
	\label{table:iou-smooth-l1}
\end{table*} 

\begin{table}[tb!]
    \centering
    \caption{Ablative study by accuracy (\%) of IoU-Smooth L1 loss  on the OBB task of DOTA-v1.0, DOTA-v1.5 and DOTA-v2.0.}
    \label{tab:baseline}
    \resizebox{0.48\textwidth}{!}{
    \begin{tabular}{c|c|c|c|c}
        \hline
        Method & Loss & DOTA-v1.0 & DOTA-v1.5 & DOTA-v2.0\\ 
        \hline
        \multirow{3}{*}{RetinaNet-H} & Smooth L1 (Reg.) & 64.17 & 56.10 & 43.06 \\
        & Smooth L1 (Reg.$^*$) & 65.78 & 57.17 & 43.92\\
        & IoU-Smooth L1 & \textbf{66.99} & \textbf{59.16} & \textbf{46.31} \\
        \hline
    \end{tabular}}
\end{table}

\begin{table*}
	\centering
	\caption{AP and mAP (\%) across categories of OBB and HBB task on DOTA. MS indicates multi-scale training and testing.}
	\resizebox{0.98\textwidth}{!}{
		\begin{tabular}{l|c|c|c|c|c|c|c|c|c|c|c|c|c|c|c|c|c}
			\hline
			\textbf{OBB Task} & Backbone &  PL &  BD &  BR &  GTF &  SV &  LV &  SH &  TC &  BC &  ST &  SBF &  RA &  HA &  SP &  HC &  mAP\\
			\hline
			\textbf{Two-stage methods} & \multicolumn{16}{|c}{} \\
			\hline
			FR-O \cite{xia2018dota} & ResNet101 \cite{he2016deep} & 79.09 & 69.12 & 17.17 & 63.49 & 34.20 & 37.16 & 36.20 & 89.19 & 69.60 & 58.96 & 49.4 & 52.52 & 46.69 & 44.80 & 46.30 & 52.93 \\
			R-DFPN \cite{yang2018automatic} & ResNet101 & 80.92 & 65.82 & 33.77 & 58.94 & 55.77 & 50.94 & 54.78 & 90.33 & 66.34 & 68.66 & 48.73 & 51.76 & 55.10 & 51.32 & 35.88 & 57.94 \\
			R$^2$CNN \cite{jiang2017r2cnn} & ResNet101 & 80.94 & 65.67 & 35.34 & 67.44 & 59.92 & 50.91 & 55.81 & 90.67 & 66.92 & 72.39 & 55.06 & 52.23 & 55.14 & 53.35 & 48.22 & 60.67 \\
			RRPN \cite{ma2018arbitrary} & ResNet101 & 88.52 & 71.20 & 31.66 & 59.30 & 51.85 & 56.19 & 57.25 & 90.81 & 72.84 & 67.38 & 56.69 & 52.84 & 53.08 & 51.94 & 53.58 & 61.01 \\
			ICN \cite{azimi2018towards} & ResNet101 & 81.40 & 74.30 & 47.70 & 70.30 & 64.90 & 67.80 & 70.00 & 90.80 & 79.10 & 78.20 & 53.60 & 62.90 & 67.00 & 64.20 & 50.20 & 68.20 \\
			RADet \cite{li2020radet} & ResNeXt101 \cite{xie2017aggregated} & 79.45 & 76.99 & 48.05 & 65.83 & 65.46 & 74.40 & 68.86 & 89.70 & 78.14 & 74.97 & 49.92 & 64.63 & 66.14 & 71.58 & 62.16 & 69.09 \\
			RoI-Transformer \cite{ding2018learning} & ResNet101 & 88.64 & 78.52 & 43.44 & 75.92 & 68.81 & 73.68 & 83.59 & 90.74 & 77.27 & 81.46 & 58.39 & 53.54 & 62.83 & 58.93 & 47.67 & 69.56 \\
			CAD-Net \cite{zhang2019cad} & ResNet101 & 87.8 & 82.4 & 49.4 & 73.5 & 71.1 & 63.5 & 76.7 & \textbf{90.9} & 79.2 & 73.3 & 48.4 & 60.9 & 62.0 & 67.0 & 62.2 & 69.9 \\
			SCRDet \cite{yang2019scrdet} & ResNet101 & 89.98 & 80.65 & 52.09 & 68.36 & 68.36 & 60.32 & 72.41 & 90.85 & {\bf87.94} & 86.86 & 65.02 & 66.68 & 66.25 & 68.24 & 65.21 & 72.61\\	
			SARD \cite{wang2019sard} & ResNet101 & 89.93 & 84.11 & 54.19 & 72.04 & 68.41 & 61.18 & 66.00 & 90.82 & 87.79 & 86.59 & 65.65 & 64.04 & 66.68 & 68.84 & 68.03 & 72.95 \\
			FADet \cite{li2019feature} & ResNet101 & 90.21 & 79.58 & 45.49 & 76.41 & 73.18 & 68.27 & 79.56 & 90.83 & 83.40 & 84.68 & 53.40 & 65.42 & 74.17 & 69.69 & 64.86 & 73.28\\
			MFIAR-Net \cite{yang2020multi} & ResNet152 \cite{he2016deep} & 89.62 & 84.03 & 52.41 & 70.30 & 70.13 & 67.64 & 77.81 & 90.85 & 85.40 & 86.22 & 63.21 & 64.14 & 68.31 & 70.21 & 62.11 & 73.49 \\
			Gliding Vertex \cite{xu2020gliding} & ResNet101 & 89.64 & 85.00 & 52.26 & \textbf{77.34} & 73.01 & 73.14 & \textbf{86.82} & 90.74 & 79.02 & 86.81 & 59.55 & \textbf{70.91} & 72.94 & 70.86 & 57.32 & 75.02 \\
			Mask OBB \cite{wang2019mask} & ResNeXt101 & 89.56 & \textbf{85.95} & 54.21 & 72.90 & 76.52 & 74.16 & 85.63 & 89.85 & 83.81 & 86.48 & 54.89 & 69.64 & 73.94 & 69.06 & 63.32 & 75.33 \\
			FFA \cite{fu2020rotation} & ResNet101 & 90.1 & 82.7 & 54.2 & 75.2 & 71.0 & 79.9 & 83.5 & 90.7 & 83.9 & 84.6 & 61.2 & 68.0 & 70.7 & 76.0 & 63.7 & 75.7 \\
			APE \cite{zhu2020adaptive} & ResNeXt-101 & 89.96 & 83.62 & 53.42 & 76.03 & 74.01 & 77.16 & 79.45 & 90.83 & 87.15 & 84.51 & 67.72 & 60.33 & 74.61 & \textbf{71.84} & 65.55 & 75.75 \\
			CSL \cite{yang2020arbitrary} & ResNet152 & \textbf{90.25} & 85.53 & 54.64 & 75.31 & 70.44 & 73.51 & 77.62 & 90.84 & 86.15 & 86.69 & 69.60 & 68.04 & 73.83 & 71.10 & 68.93 & 76.17\\
			SCRDet++ (FPN) & ResNet101 & 89.77 & 83.90 & \textbf{56.30} & 73.98 & 72.60 & 75.63 & 82.82 & 90.76 & 87.89 & 86.14 & 65.24 & 63.17 & \textbf{76.05} & 68.06 & 70.24 & 76.20 \\
			SCRDet++ MS (FPN) & ResNet101 & 90.05 & 84.39 & 55.44 & 73.99 & \textbf{77.54} & 71.11 & 86.05 & 90.67 & 87.32 & \textbf{87.08} & \textbf{69.62} & 68.90 & 73.74 & 71.29 & 65.08 & \textbf{76.81} \\
% 			SCRDet++ MS (FPN-based) & ResNet152 & 90.06 & 84.37 & 56.76 & 75.82 & 79.33 & 74.50 & 86.51 & 90.77 & 88.51 & 86.84 & 69.60 & 69.05 & 76.15 & 79.92 & 76.61 & 78.99\\
			\hline
			\textbf{Single-stage methods} & \multicolumn{16}{|c}{} \\
			\hline
			IENet \cite{lin2019ienet} & ResNet101 & 80.20 & 64.54 & 39.82 & 32.07 & 49.71 & 65.01 & 52.58 & 81.45 & 44.66 & 78.51 & 46.54 & 56.73 & 64.40 & 64.24 & 36.75 & 57.14 \\
			Axis Learning \cite{xiao2020axis} & ResNet101 & 79.53 & 77.15 & 38.59 & 61.15 & 67.53 & 70.49 & 76.30 & 89.66 & 79.07 & 83.53 & 47.27 & 61.01 & 56.28 & 66.06 & 36.05 & 65.98\\
			P-RSDet \cite{zhou2020arbitrary} & ResNet101 & 89.02 & 73.65 & 47.33 & 72.03 & 70.58 & 73.71 & 72.76 & 90.82 & 80.12 & 81.32 & 59.45 & 57.87 & 60.79 & 65.21 & 52.59 & 69.82 \\
			O$^2$-DNet \cite{wei2020oriented} & Hourglass104 \cite{newell2016stacked} & 89.31 & 82.14 & 47.33 & 61.21 & 71.32 & 74.03 & 78.62 & 90.76 & 82.23 & 81.36 & 60.93 & 60.17 & 58.21 & 66.98 & 61.03 & 71.04 \\
			R$^3$Det \cite{ yang2021r3det} & ResNet152 & 89.24 & 80.81 & 51.11 & 65.62 & 70.67 & 76.03 & 78.32 & 90.83 & 84.89 & 84.42 & 65.10 & 57.18 & 68.10 & 68.98 & 60.88 & 72.81\\
			RSDet \cite{qian2021learning} & ResNet152 & 90.1 & 82.0 & 53.8 & 68.5 & 70.2 & 78.7 & 73.6 & 91.2 & 87.1 & 84.7 & 64.3 & 68.2 & 66.1 & 69.3 & 63.7 & 74.1 \\ 
			SCRDet++ (R$^3$Det) & ResNet152 & 89.20 & 83.36 & 50.92 & 68.17 & 71.61 & 80.23 & 78.53 & 90.83 & 86.09 & 84.04 & 65.93 & 60.8 & 68.83 & 71.31 & 66.24 & 74.41 \\
			SCRDet++ MS (R$^3$Det) & ResNet152 & 88.68 & 85.22 & 54.70 & 73.71 & 71.92 & \textbf{84.14} & 79.39 & 90.82 & 87.04 & 86.02 & 67.90 & 60.86 & 74.52 & 70.76 & \textbf{72.66} & 76.56\\
			\hline
			
			~&\multicolumn{16}{c}{}\\
			
			\hline
			\textbf{HBB Task} & Backbone &  PL &  BD &  BR &  GTF &  SV &  LV &  SH &  TC &  BC &  ST &  SBF &  RA &  HA &  SP &  HC &  mAP\\
			\hline
			\textbf{Two-stage methods} & \multicolumn{16}{|c}{} \\
			\hline
			FR-H \cite{ren2017faster} & ResNet101 & 80.32 & 77.55 & 32.86 & 68.13 & 53.66 & 52.49 & 50.04 & 90.41 & 75.05 & 59.59 & 57.00 & 49.81 & 61.69 & 56.46 & 41.85 & 60.46 \\
			ICN \cite{azimi2018towards} & ResNet101 & 90.00 & 77.70 & 53.40 & 73.30 & 73.50 & 65.00 & 78.20 & 90.80 & 79.10 & 84.80 & 57.20 & 62.10 & 73.50 & 70.20 & 58.10 & 72.50 \\
			IoU-Adapt R-CNN \cite{yan2019iou} & ResNet101 & 88.62 & 80.22 & 53.18 & 66.97 & 76.30 & 72.59 & 84.07 & 90.66 & 80.95 & 76.24 & 57.12 & 66.65 & 84.08 & 66.36 & 56.85 & 72.72 \\
			SCRDet \cite{yang2019scrdet} & ResNet101 & {\bf 90.18} & 81.88 & 55.30 & 73.29 & 72.09 & 77.65 & 78.06 & {\bf 90.91} & 82.44 & 86.39 & 64.53 & 63.45 & 75.77 & 78.21 & 60.11 & 75.35 \\
			FADet \cite{li2019feature} & ResNet101 & 90.15 & 78.60 & 51.92 & \textbf{75.23} & 73.60 & 71.27 & 81.41 & 90.85 & 83.94 & 84.77 & 58.91 & 65.65 & 76.92 & 79.36 & 68.17 & 75.38\\
			Mask OBB \cite{wang2019mask} & ResNeXt-101 & 89.69 & \textbf{87.07} & 58.51 & 72.04 & 78.21 & 71.47 & 85.20 & 89.55 & 84.71 & 86.76 & 54.38 & 70.21 & 78.98 & 77.46 & 70.40 & 76.98 \\
			A$^2$RMNet \cite{qiu2019a2rmnet} & ResNet101 & 89.84 & 83.39 & 60.06 & 73.46 & \textbf{79.25} & 83.07 & \textbf{87.88} & 90.90 & 87.02 & \textbf{87.35} & 60.74 & 69.05 & 79.88 & 79.74 & 65.17 & 78.45\\
			SCRDet++ (FPN) & ResNet101 & 90.01 & 82.32 & 61.94 & 68.62 & 69.62 & 81.17 & 78.83 & 90.86 & 86.32 & 85.10 & 65.10 & 61.12 & 77.69 & \textbf{80.68} & 64.25 & 76.24 \\
			SCRDet++ MS (FPN) & ResNet101 & 90.00 & 86.25 & \textbf{65.04} & 74.52 & 72.93 & \textbf{84.17} & 79.05 & 90.72 & \textbf{87.37} & 87.06 & \textbf{72.10} & 66.72 & \textbf{82.64} & 80.57 & \textbf{71.07} & \textbf{79.35} \\
			\hline
			\textbf{Single-stage methods} & \multicolumn{16}{|c}{} \\
			\hline
			SBL \cite{sun2018salience} & ResNet50 & 89.15 & 66.04 & 46.79 & 52.56 & 73.06 & 66.13 & 78.66 & 90.85 & 67.40 & 72.22 & 39.88 & 56.89 & 69.58 & 67.73 & 34.74 & 64.77 \\
			FMSSD \cite{wang2019fmssd} & VGG16 \cite{simonyan2015very} & 89.11 & 81.51 & 48.22 & 67.94 & 69.23 & 73.56 & 76.87 & 90.71 & 82.67 & 73.33 & 52.65 & \textbf{67.52} & 72.37 & 80.57 & 60.15 & 72.43 \\
			EFR \cite{fu2019enhanced} & VGG16 & 88.36 & 83.90 & 45.78 & 67.24 & 76.80 & 77.15 & 85.35 & 90.77 & 85.55 & 75.77 & 54.64 & 60.76 & 71.40 & 77.90 & 60.94 & 73.49 \\
			SCRDet++ (RetinaNet) & ResNet152 & 87.89 & 84.64 & 56.94 & 68.03 & 74.67 & 78.75 & 78.50 & 90.80 & 85.60 & 84.98 & 53.56 & 56.75 & 76.66 & 75.08 & 62.75 & 74.37 \\
			\hline
	\end{tabular}}
	\label{table:dota_sota}
\end{table*}

\begin{table*}
	\centering
	\caption{Accuracy (\%) on DIOR.  $^*$ indicates our own implementation, higher than the official baseline. $^\dagger$ indicates data augmentation is used.}
	\resizebox{1.0\textwidth}{!}{
		\begin{tabular}{l|c|c|c|c|c|c|c|c|c|c|c|c|c|c|c|c|c|c|c|c|c|c}
			\hline
			& Backbone &  c1 &  c2 &  c3 &  c4 &  c5 &  c6 &  c7 &  c8 &  c9 &  c10 &  c11 &  c12 &  c13 &  c14 &  c15 & c16 & c17 & c18 & c19 & c20 &  mAP\\
			\hline
			\textbf{Two-stage methods} & \multicolumn{21}{|c}{} \\
			\hline
			Faster-RCNN \cite{ren2017faster} & VGG16 & 53.6 & 49.3 & 78.8 & 66.2 & 28.0 & 70.9 & 62.3 & 69.0 & 55.2 & 68.0 & 56.9 & 50.2 & 50.1 & 27.7 & 73.0 & 39.8 & 75.2 & 38.6 & 23.6 & 45.4 & 54.1 \\
			\hline
			\multirow{2}{*}{Mask‐RCNN \cite{he2017mask} }
			& ResNet‐50 & 53.8 & 72.3 & 63.2 & 81.0 & 38.7 & 72.6 & 55.9 & 71.6 & 67.0 & 73.0 & 75.8 & 44.2 & 56.5 & 71.9 & 58.6 & 53.6 & 81.1 & 54.0 & 43.1 & 81.1 & 63.5\\
			& ResNet‐101 & 53.9 & 76.6 & 63.2 & 80.9 & 40.2 & 72.5 & 60.4 & 76.3 & 62.5 & 76.0 & 75.9 & 46.5 & 57.4 & 71.8 & 68.3 & 53.7 & 81.0 & 62.3 & 43.0 & 81.0 & 65.2 \\
			\hline
			\multirow{2}{*}{PANet \cite{liu2018path} }
			& ResNet‐50 & 61.9 & 70.4 & 71.0 & 80.4 & 38.9 & 72.5 & 56.6 & 68.4 & 60.0 & 69.0 & 74.6 & 41.6 & 55.8 & 71.7 & 72.9 & 62.3 & 81.2 & 54.6 & 48.2 & 86.7 & 63.8\\
			& ResNet‐101 & 60.2 & 72.0 & 70.6 & 80.5 & 43.6 & 72.3 & 61.4 & 72.1 & 66.7 & 72.0 & 73.4 & 45.3 & 56.9 & 71.7 & 70.4 & 62.0 & 80.9 & 57.0 & 47.2 & 84.5 & 66.1 \\
			\hline
			CornerNet \cite{law2018cornernet} & Hourglass104 & 58.8 & 84.2 & 72.0 & 80.8 & 46.4 & 75.3 & 64.3 & 81.6 & 76.3 & 79.5 & 79.5 & 26.1 & 60.6 & 37.6 & 70.7 & 45.2 & 84.0 & 57.1 & 43.0 & 75.9 & 64.9 \\
			\hline
			\multirow{2}{*}{FPN \cite{lin2017feature}}
			& ResNet‐50 & 54.1 & 71.4 & 63.3 & 81.0 & 42.6 & 72.5 & 57.5 & 68.7 & 62.1 & 73.1 & 76.5 & 42.8 & 56.0 & 71.8 & 57.0 & 53.5 & 81.2 & 53.0 & 43.1 & 80.9 & 63.1\\
			& ResNet‐101 & 54.0 & 74.5 & 63.3 & 80.7 & 44.8 & 72.5 & 60.0 & 75.6 & 62.3 & 76.0 & 76.8 & 46.4 & 57.2 & 71.8 & 68.3 & 53.8 & 81.1 & 59.5 & 43.1 & 81.2 & 65.1 \\
			\hline
			CSFF \cite{cheng2020cross} & ResNet-101 & 57.2 & 79.6 & 70.1 & 87.4 & 46.1 & 76.6 & 62.7 & 82.6 & 73.2 & 78.2 & 81.6 & 50.7 & 59.5 & \textbf{73.3} & 63.4 & 58.5 & 85.9 & 61.9 & 42.9 & 86.9 & 68.0\\
			\hline
			FPN$^*$ & ResNet‐50 & 66.57 & 83.00 & 71.89 & 83.02 & 50.41 & 75.74 & 70.23 & 81.08 & 74.83 & 79.03 & 77.74 & 55.29 & 62.06 & 72.26 & 72.10 & 68.64 & 81.20 & 66.07 & 54.56 & 89.09 & 71.74 \\
			SCRDet++ (FPN$^*$) & ResNet‐50 & 66.35 & 83.36 &74.34 & 87.33 & 52.45 & 77.98 & 70.06 & 84.22 & 77.95 & 80.73 & 81.26 & 56.77 & 63.70 & 73.29 & 71.94 & \textbf{71.24} & 83.40 & 62.28 & 55.63 & 90.00 & 73.21 \\
			SCRDet++ (FPN$^*$)$^\dagger$ & ResNet‐101 & \textbf{80.79} & \textbf{87.67} & \textbf{80.46} & \textbf{89.76} & \textbf{57.83} & \textbf{80.90} & 75.23 & \textbf{90.01} & \textbf{82.93} & \textbf{84.51} & \textbf{83.55} & 63.19 & \textbf{67.25} & 72.59 & \textbf{79.20} & 70.44 & \textbf{89.97} & 70.71 & \textbf{58.82} & \textbf{90.25} & \textbf{77.80} \\
			\hline
			\textbf{Single-stage methods} & \multicolumn{21}{|c}{} \\
			\hline
			SSD \cite{liu2016ssd} & VGG16 & 59.5 & 72.7 & 72.4 & 75.7 & 29.7 & 65.8 & 56.6 & 63.5 & 53.1 & 65.3 & 68.6 & 49.4 & 48.1 & 59.2 & 61.0 & 46.6 & 76.3 & 55.1 & 27.4 & 65.7 & 58.6 \\
			YOLOv3 \cite{redmon2018yolov3} & Darknet‐53 & 72.2 & 29.2 & 74.0 & 78.6 & 31.2 & 69.7 & 26.9 & 48.6 & 54.4 & 31.1 & 61.1 & 44.9 & 49.7 & 87.4 & 70.6 & 68.7 & 87.3 & 29.4 & 48. 3 & 78.7 & 57.1 \\
			\hline
			\multirow{2}{*}{RetinaNet \cite{lin2017focal}}
			& ResNet‐50 & 53.7 & 77.3 & 69.0 & 81.3 & 44.1 & 72.3 & 62.5 & 76.2 & 66.0 & 77.7 & 74.2 & 50.7 & 59.6 & 71.2 & 69.3 & 44.8 & 81.3 & 54.2 & 45.1 & 83.4 & 65.7\\
			& ResNet‐101 & 53.3 & 77.0 & 69.3 & 85.0 & 44.1 & 73.2 & 62.4 & 78.6 & 62.8 & 78.6 & 76.6 & 49.9 & 59.6 & 71.1 & 68.4 & 45.8 & 81.3 & 55.2 & 44.4 & 85.5 & 66.1 \\
			\hline
			RetinaNet$^*$ & ResNet‐50 & 59.98 & 79.02 & 70.85 & 83.37 & 45.25 & 75.93 & 64.53 & 76.87 & 66.63 & 80.25 & 76.75 & 55.94 & 60.70 & 70.38 & 61.45 & 60.15 & 81.13 & 62.76 & 44.52 & 84.46 & 68.05 \\
			SCRDet++ (RetinaNet$^*$) & ResNet‐50 & 64.33 & 78.99 & 73.24 & 85.72 & 45.83 & 75.99 & 68.41 & 79.28 & 68.93 & 77.68 & 77.87 & 56.70 & 62.15 & 70.38 & 67.66 & 60.42 & 80.93 & 63.74 & 44.44 & 84.56 & 69.36 \\
			SCRDet++ (RetinaNet$^*$)$^\dagger$ & ResNet‐101 & 71.94 & 84.99 & 79.48 & 88.86 & 52.27 & 79.12 & \textbf{77.63} & 89.52 & 77.79 & 84.24 & 83.07 & \textbf{64.22} & 65.57 & 71.25 & 76.51 & 64.54 & 88.02 & \textbf{70.91} & 47.12 & 85.10 & 75.11 \\
			\hline
	\end{tabular}}	
	\label{table:dior_sota}
%	\vspace{-8pt}
\end{table*}

\subsection{Datasets and Protocols}\label{subsec:dataset}
We choose a wide variety of public datasets from both aerial images as well as natural images and scene texts for evaluation. The details are as follows.

\textbf{DOTA} \cite{xia2018dota}: DOTA-v1.0 is a complex aerial image dataset for object detection, which contains objects exhibiting a wide variety of scales, orientations, and shapes. DOTA-v1.0 contains 2,806 aerial images and 15 common object categories from different sensors and platforms. The fully annotated DOTA-v1.0 benchmark contains 188,282 instances, each of which is labeled by an arbitrary quadrilateral. There are two detection tasks for DOTA: horizontal bounding boxes (HBB) and oriented bounding boxes (OBB). The training set, validation set, and test set account for 1/2, 1/6, 1/3 of the entire data set, respectively. 
In contrast, DOTA-v1.5 uses the same images as DOTA-v1.0, but extremely small instances (less than 10 pixels) are also annotated. Moreover, a new category, containing 402,089 instances in total is added in this version. While DOTA-v2.0 contains 18 common categories, 11,268 images and 1,793,658 instances. Compared to DOTA-v1.5, it includes the new categories. The 11,268 images in DOTA-v2.0 are split into training, validation, test-dev, and test-challenge sets.
We divide the images into $ 600 \times 600 $ subimages with an overlap of 150 pixels and scale it to $ 800 \times 800 $. 
% With all these processes, we obtain about 27,000 patches. The model is trained by 135k iterations in total, and the learning rate changes during the 81k and 108k iterations from 5e-4 to 5e-6. 
The short names for categories are defined as (abbreviation-full name): PL-Plane, BD-Baseball diamond, BR-Bridge, GTF-Ground field track, SV-Small vehicle, LV-Large vehicle, SH-Ship, TC-Tennis court, BC-Basketball court, ST-Storage tank, SBF-Soccer-ball field, RA-Roundabout, HA-Harbor, SP-Swimming pool, HC-Helicopter, CC-container crane, AP-airport and HP-helipad.

\textbf{DIOR} \cite{li2020object}: DIOR is another large aerial images dataset labeled by a horizontal bounding box. It consists of 23,463 images and 190,288 instances, covering 20 object classes. DIOR has a large variation of object size, not only in spatial resolutions, but also in the aspect of inter‐class and intra‐class size variability across objects. The complexity of DIOR is also reflected in different imaging conditions, weathers, seasons, and image quality, and it has high inter‐class similarity and intra‐class diversity. The training protocol of DIOR is basically consistent with DOTA-v1.0. The short names c1-c20 for categories in our experiment are defined as: Airplane, Airport, Baseball field, Basketball court, Bridge, Chimney, Dam, Expressway service area, Expressway toll station, Golf field, Ground track field, Harbor, Overpass, Ship, Stadium, Storage tank, Tennis court, Train station, Vehicle, and Wind mill.

\textbf{UCAS-AOD} \cite{zhu2015orientation}: UCAS-AOD contains 1,510 aerial images of approximately 659 $ \times $ 1,280 pixels, it contains two categories of 14,596 instances. In line with \cite{xia2018dota, azimi2018towards}, we randomly select 1,110 for training and 400 for testing.

\textbf{BSTLD} \cite{behrendt2017deep}: BSTLD contains 13,427 camera images at a resolution of 720 $ \times $ 1,280 pixels and contains about 24,000 annotated small traffic lights. Specifically, 5,093 training images are annotated by 15 labels every 2 seconds, but only 3,153 images contain the instance, about 10,756. There are very few instances of many categories, so we reclassify them into 4 categories (red, yellow, green, off). In contrast, 8,334 consecutive test images are annotated by 4 labels at about 15 fps. In this paper, we only use the training set of BSTLD, whose median traffic light width is 8.6 pixels. In the experiment, we divide BSTLD training set into a training set and a test set according to the ratio of $6:4$. Note that we use the RetinaNet with P2 feature level and FPN to verify InLD, and scale the size of the input image to 720 $\times$ 1,280.

\textbf{S$^2$TLD}: S$^2$TLD\footnote{S$^2$TLD is available at \url{https://github.com/Thinklab-SJTU/S2TLD}} is our collected and annotated traffic light dataset as released in this paper. It contains 5,786 images of approximately 1,080 $\times$ 1,920 pixels (1,222 images) and 720 $\times$ 1,280 pixels (4,564 images). It also contains 5 categories (namely red, yellow, green, off and wait on) of 14,130 instances. The scenes cover a variety of lighting, weather and traffic conditions, including busy street scenes inner-city, dense stop-and-go traffic, strong changes in illumination/exposure, flickering/fluctuating traffic lights, multiple visible traffic lights, image parts that can be confused with traffic lights (e.g. large round tail lights), as shown in Fig. \ref{fig:S2TLD_VIS}. The training strategy is consistent with BSTLD.

In addition to the above datasets, we also use natural image dataset COCO \cite{lin2014microsoft} and scene text dataset ICDAR2015 \cite{karatzas2015icdar} for further evaluation. 

%By standard practices, we adopt COCO train2017 for training and conducte ablation studies on the remaining 5k validation images (minival). For ICDAR2015 dataset, 1000 of which are used for training and the remaining are for testing. The ICDAR2015 dataset uses the same learning strategy and changes the learning rate size in 15k iterations, 20k iterations, and 25k iterations, respectively.

The experiments are initialized by ResNet50 \cite{he2016deep} by default unless otherwise specified. The weight decay and momentum for all experiments are set 0.0001 and 0.9, respectively. We employ MomentumOptimizer over 8 GPUs with a total of 8 images per minibatch.  We follow the standard evaluation protocol of COCO, while for other datasets, the anchors of RetinaNet-based method have areas of $32^{2}$ to $512^2$ on pyramid levels from P3 to P7, respectively. At each pyramid level we use anchors at seven aspect ratios $\{1,1/2,2,1/3,3,5,1/5\}$ and three scales $\{2^{0}, 2^{1/3}, 2^{2/3}\}$. For rotating
anchor-based method (RetinaNet-R), the angle is set by an arithmetic progression from $-90^\circ$ to $-15^\circ$ with an interval of $15$ degrees.

\subsection{Ablation Study}
The ablation study covers the detailed evaluation of the effect of image level denoising (ImLD) and instance level denoising (InLD), as well as their combination.

\textbf{Effect of Image-Level Denoising.}
We experiment with five denoising modules introduced in \cite{xie2019feature} on DOTA-v1.0. We use our previous work R$^3$Det \cite{ yang2021r3det}, one of the most state-of-the-art methods on the DOTA-v1.0, as the baseline. From Tab. \ref{table:ImLD_Ablation_Study}, one can observe that most methods work effectively except the mean filtering. Among them, the non-local with Gaussian is the most effective (1.95\% higher).

\begin{figure}[!tb]
	\centering
	\subfigure[COCO: the red boxes represent missed objects and the orange boxes represent false alarms.]{
		\begin{minipage}[t]{0.45\linewidth}
			\centering
			\includegraphics[width=1.0\textwidth]{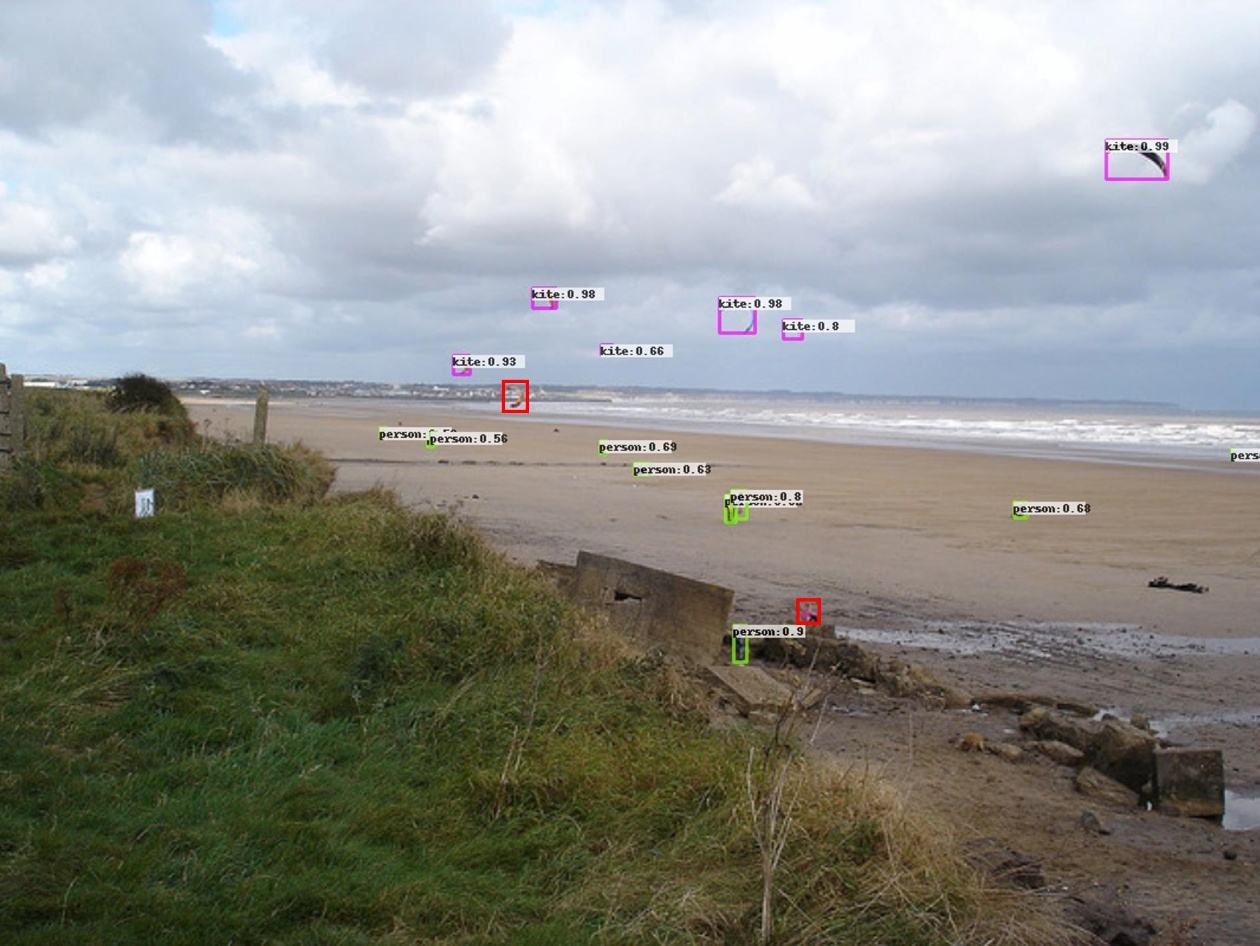}
			\centering
			\label{fig:COCO_val2014_000000224664_}
	    \end{minipage}
		\begin{minipage}[t]{0.45\linewidth}
			\centering
			\includegraphics[width=1.0\textwidth]{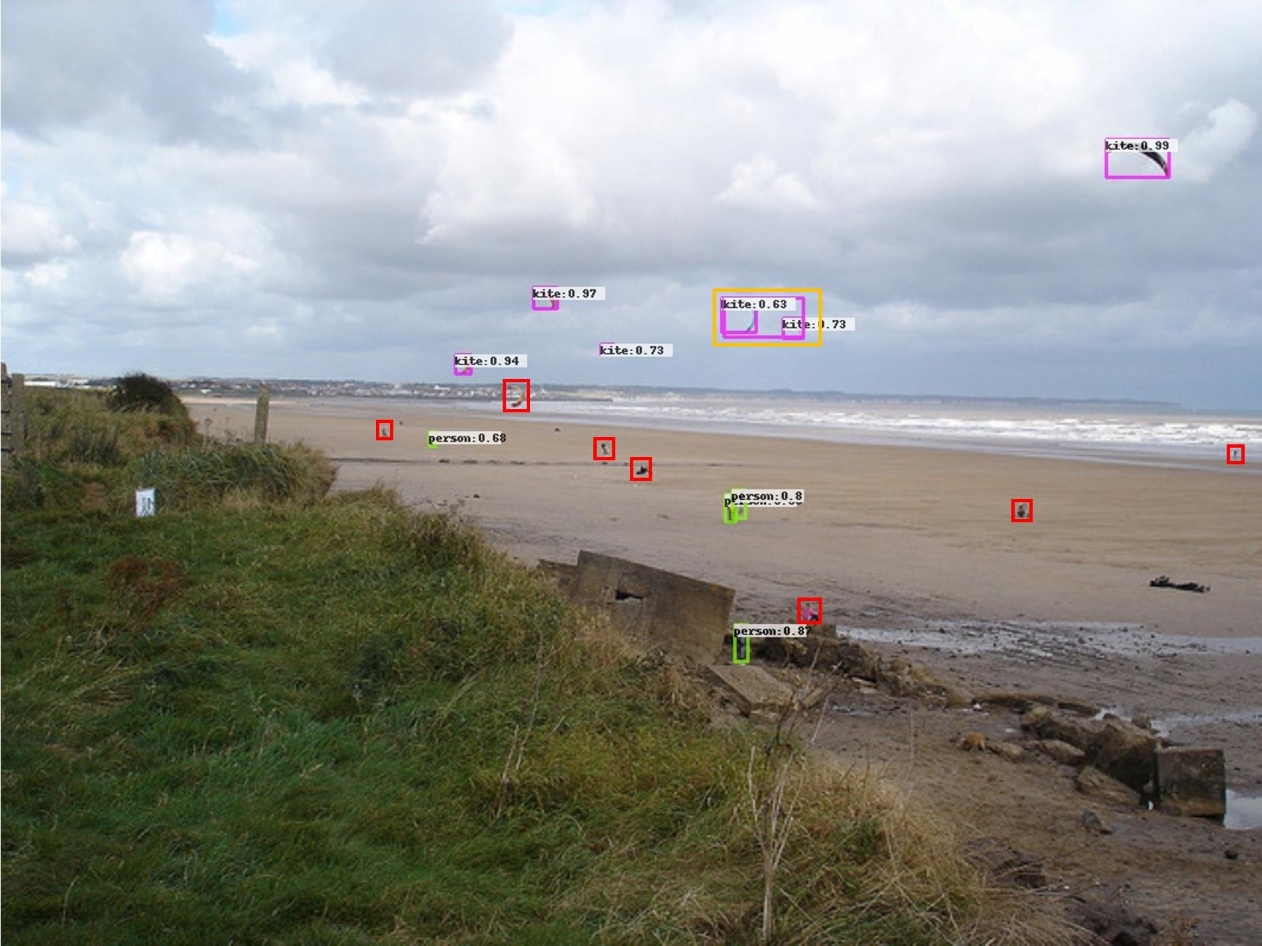}
			\centering
			\label{fig:COCO_val2014_000000224664_bs_}
	    \end{minipage}}\\
	\subfigure[ICDAR2015: red arrows denote missed objects.]{
		\begin{minipage}[t]{0.45\linewidth}
			\centering
			\includegraphics[width=1.0\textwidth]{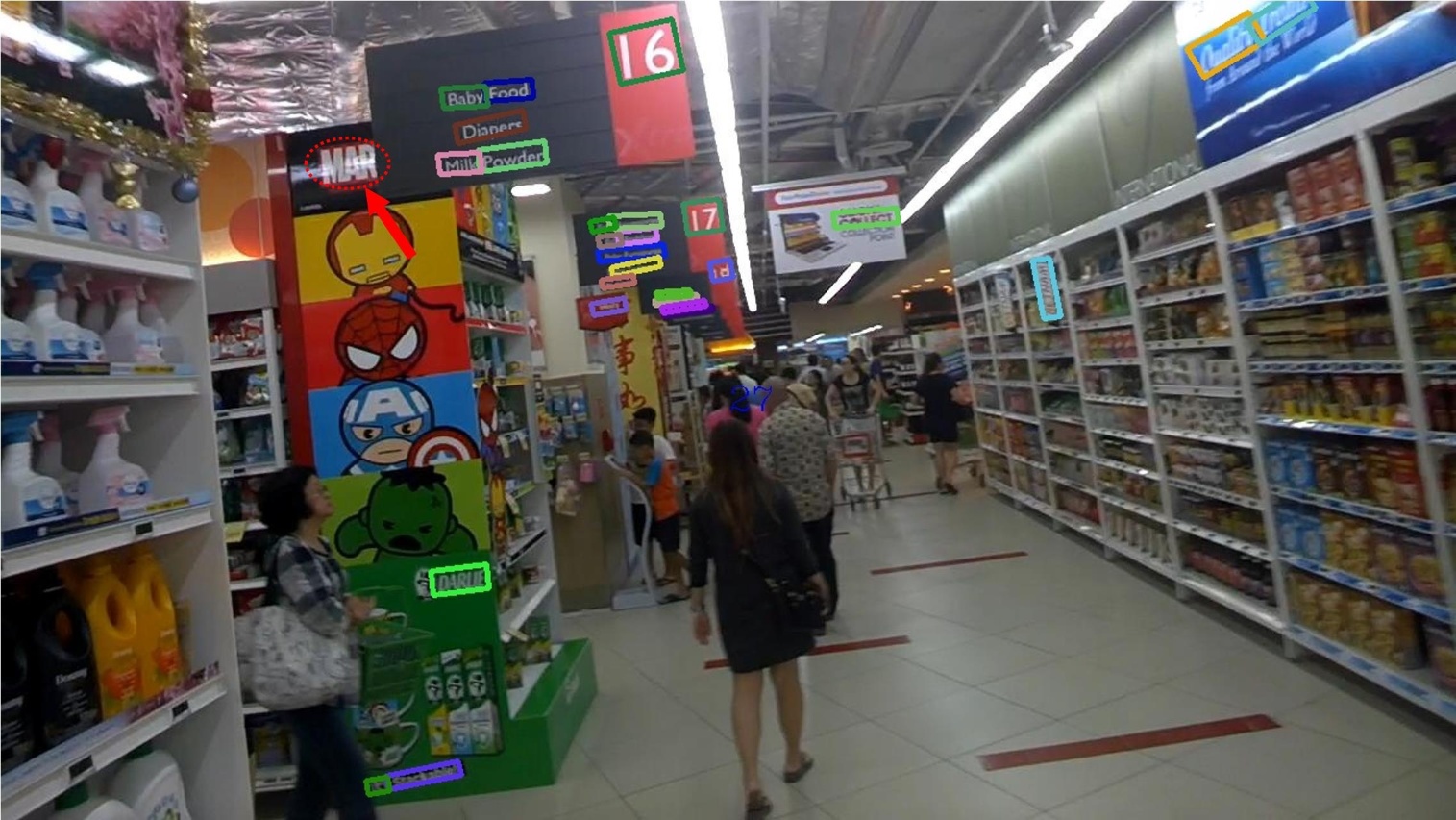}
			\centering
			\label{fig:img_108_R2CNN++}
	    \end{minipage}
	    \begin{minipage}[t]{0.45\linewidth}
    			\centering
    			\includegraphics[width=1.0\textwidth]{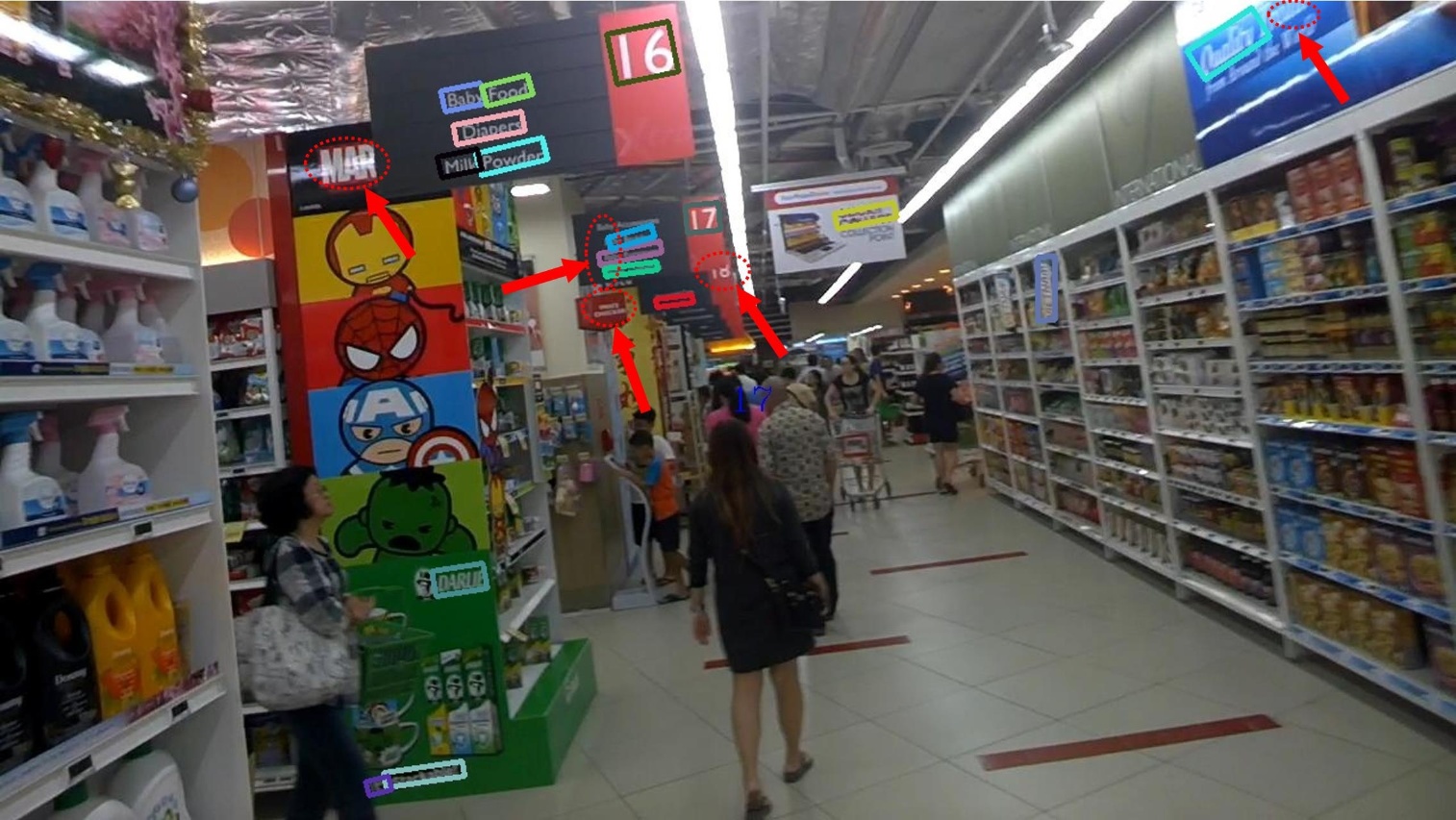}
    			\centering
    			\label{fig:img_108_R2CNN_}
    	\end{minipage}}\\
    \subfigure[S$^2$TLD: the red box represent missed object.]{
		\begin{minipage}[t]{0.45\linewidth}
			\centering
			\includegraphics[width=1.0\textwidth]{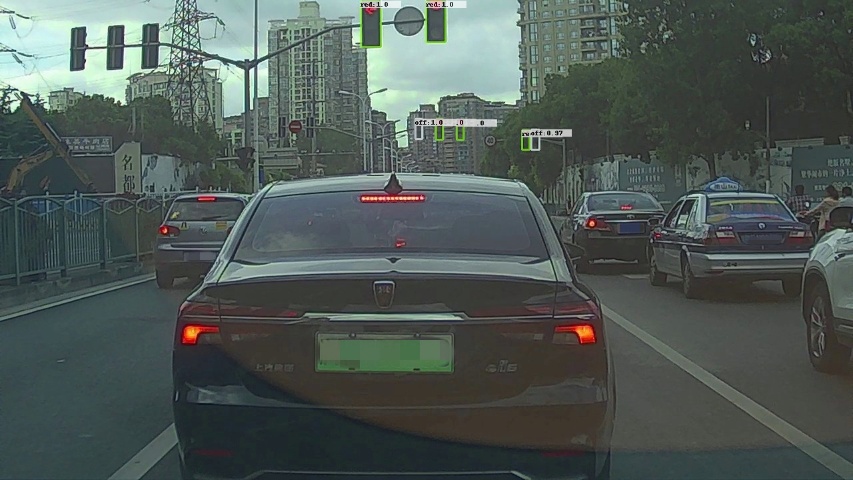}
			\centering
			\label{fig:001961_InLD}
	    \end{minipage}
	    \begin{minipage}[t]{0.45\linewidth}
    			\centering
    			\includegraphics[width=1.0\textwidth]{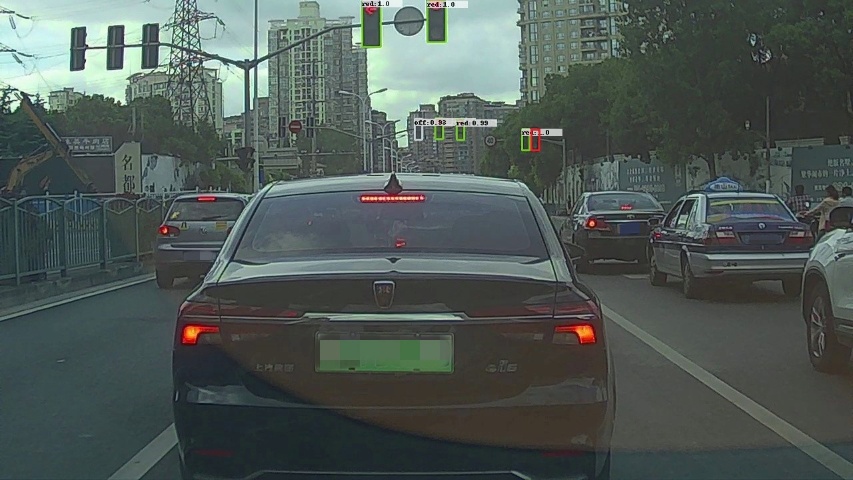}
    			\centering
    			\label{fig:001961}
    	\end{minipage}}\\
	\caption{Visual illustration of detection results on the datasets of COCO, ICDAR2015, S$^2$TLD before (right) and after (left) using InLD.}
	\label{fig:InLD_vis}
	%\vspace{-10pt}
\end{figure}

\textbf{Effect of Instance-Level Denoising.}
The purpose of designing InLD is to make the feature of different categories decoupled in the channel dimension, while the features of the object and non-object are enhanced and weakened in the spatial dimension, respectively. We have designed some verification tests and obtained positive results as shown in Tab. \ref{table:InLD_Ablative_Study}. We first explore the utility of weakening the non-object noise by binary semantic segmentation, and the detection mAP has increased from 65.73\% to 68.12\%. The result on multi-category semantic segmentation further proves that there is indeed interference between objects, which is reflected by the $1.31\%$ increase of detection mAP (reaching 69.43\%). From the above two experiments, we can preliminarily speculate that the interference in the non-object area is the main reason that affects the performance of the detector. It is surprising to find that coproducting the prediction score for objectness (see $P(object)$ in Eq.~\ref{eq:objectness}) can further improve performance and speed up training with a final accuracy of 69.81\%. Experiments in Tab. \ref{table:dota_ablation_study} show that InLD has greatly improved the R$^3$Det's performance of small objects, such as BR, SV, LV, SH, SP, HC, which increased by 3.94\%, 0.84\%, 4.32\%, 8.48\%, 10.15\%, and 9.41\%, respectively. While the accuracy is greatly improved, the detection speed of the model is only reduced by 1fps (at 13fps). In addition to the DOTA-v1.0 dataset, we have used more datasets to verify the general applicability, such as DIOR, ICDAR, COCO and S$^2$TLD. InLD obtains 1.44\%, 1.55\%, 1.4\% and 0.86\% improvements in each of the four datasets according to Tab. \ref{table:ImLD_and_InLD} and Fig. \ref{fig:InLD_vis} shows the visualization results before and after using InLD. In order to investigate whether the performance improvement brought by InLD is due to the extra computation (dilated convolutions) or supervised learning ($L_{InLD}$), we perform ablation experiments by controlling the number of dilated convolutions and supervision signal. Tab. \ref{table:InLD} shows that supervised learning is the main contribution of InLD rather than more convolution layers.

\begin{figure}[!tb]
	\centering
	\subfigure[BC and TC]{
		\begin{minipage}[t]{0.3\linewidth}
			\centering
			\includegraphics[width=1.0\textwidth, height=3.5cm]{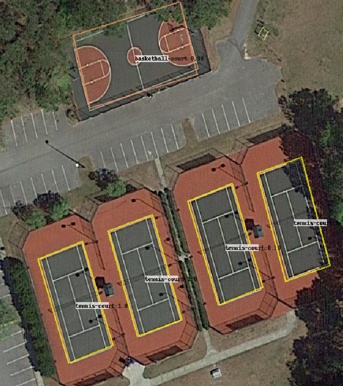}
			\centering
	\end{minipage}}
	\subfigure[SBF and GTF]{
		\begin{minipage}[t]{0.3\linewidth}
			\centering
			\includegraphics[width=1.0\textwidth, height=3.5cm]{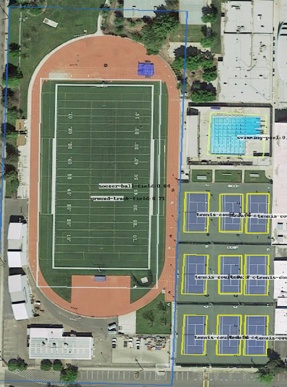}
	\end{minipage}}
	\subfigure[HA and SH]{
		\begin{minipage}[t]{0.3\linewidth}
			\centering
			\includegraphics[width=1.0\textwidth, height=3.5cm]{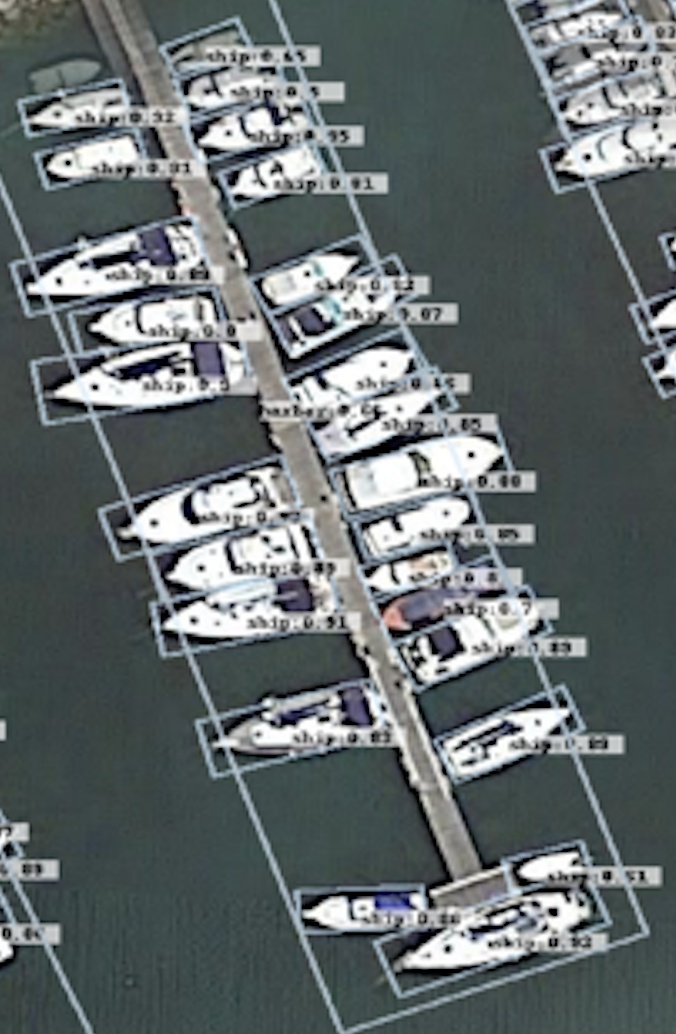}	
	\end{minipage}}\\
	%\vspace{-5pt}
	\subfigure[SP]{
		\begin{minipage}[t]{0.22\linewidth}
			\centering
			\includegraphics[width=1.0\textwidth, height=2.5cm]{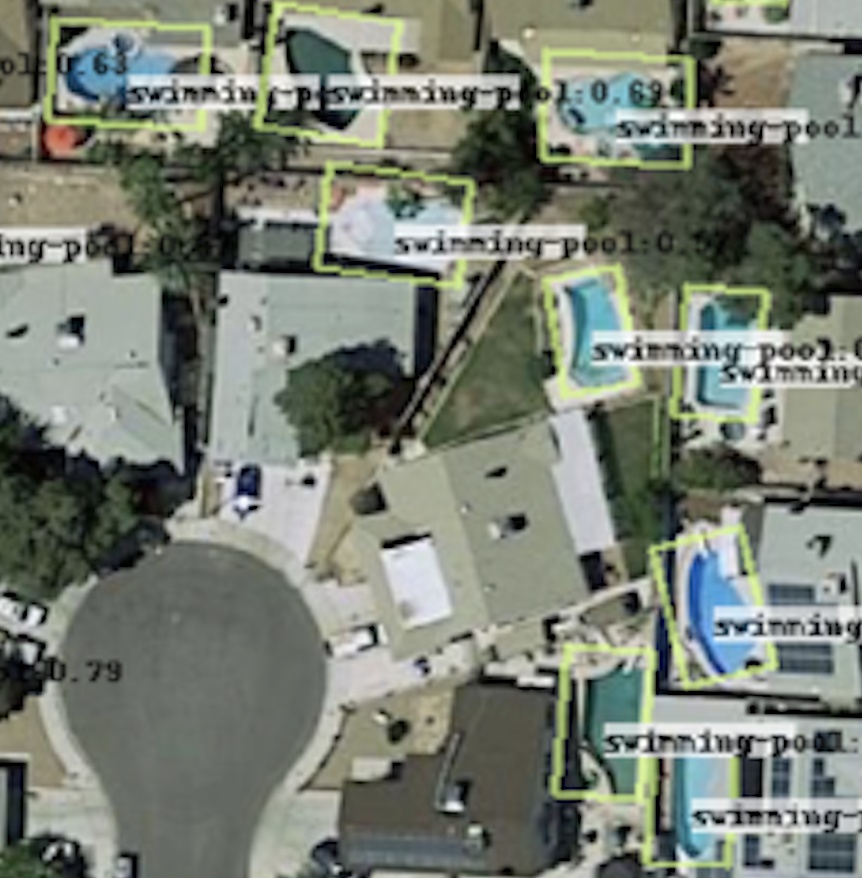}
			\centering
	\end{minipage}}
	\subfigure[RA and SV]{
		\begin{minipage}[t]{0.22\linewidth}
			\centering
			\includegraphics[width=1.0\textwidth, height=2.5cm]{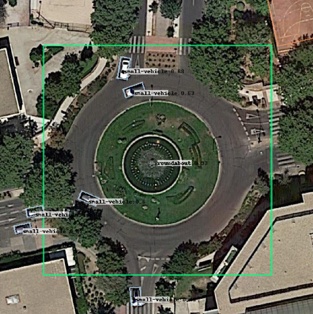}
			\centering
	\end{minipage}}
	\subfigure[ST]{
		\begin{minipage}[t]{0.22\linewidth}
			\centering
			\includegraphics[width=1.0\textwidth, height=2.5cm]{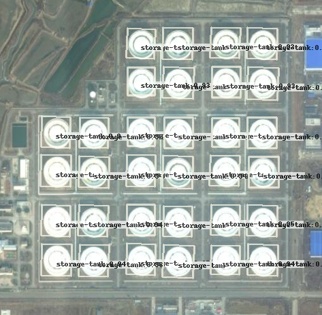}
			\centering
	\end{minipage}}
	\subfigure[BD and RA]{
		\begin{minipage}[t]{0.22\linewidth}
			\centering
			\includegraphics[width=1.0\textwidth, height=2.5cm]{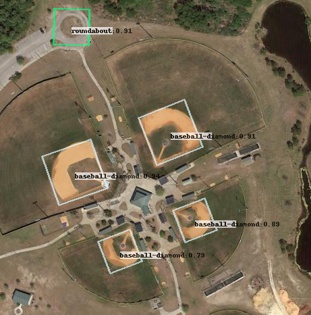}
			\centering
	\end{minipage}}\\
	%\vspace{-5pt}
	\subfigure[SV and LV]{
		\begin{minipage}[t]{0.3\linewidth}
			\centering
			\includegraphics[width=1.0\textwidth, height=3cm]{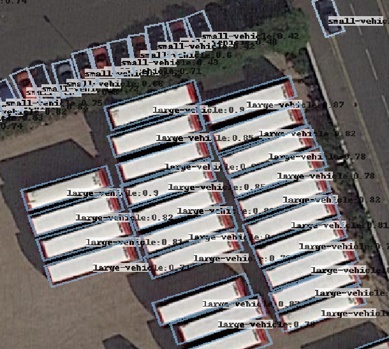}
			\centering
	\end{minipage}}
	\subfigure[PL and HC]{
		\begin{minipage}[t]{0.3\linewidth}
			\centering
			\includegraphics[width=1.0\textwidth, height=3cm]{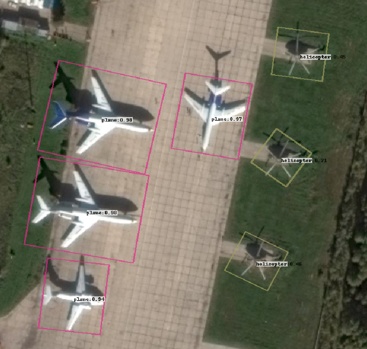}
	\end{minipage}}
	\subfigure[BR]{
		\begin{minipage}[t]{0.3\linewidth}
			\centering
			\includegraphics[width=1.0\textwidth, height=3cm]{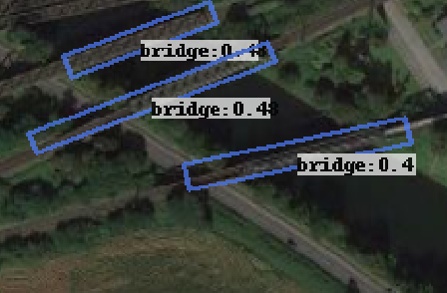}	
	\end{minipage}}
	
	\caption{Visual illustration of detection results on OBB task on DOTA-v1.0 of different objects by the proposed method.}
	\label{fig:DOTA}
	%\vspace{-5pt}
\end{figure}

In particular, we conduct a detailed study on the SJTU Small Traffic Light Dataset (S$^2$TLD) which is our newly released traffic detection dataset. Compared with BSTLD, S$^2$TLD has more available categories. In addition, S$^2$TLD contains two different resolution images taken from two different cameras, which can be used for more challenging detection tasks. Tab. \ref{table:STLD} shows the effectiveness of InLD on these two traffic light datasets.

\textbf{Effect of combining ImLD and InLD.}
A natural idea is whether we can combine these two denoising structures, as shown in Fig. \ref{fig:pipeline}. For more comprehensive study, we perform detailed ablation experiments on different datasets and different detection tasks. The experimental results are listed in Tab. \ref{table:ImLD_and_InLD}, and we tend to get the following remarks:

1) Most of the datasets are relatively clean, so ImLD does not obtain a significant increase in all datasets.

2) The performance improvement of detectors with InLD is very significant and stable, and is superior to ImLD.

3) The gain by the combination of ImLD and InLD is not large, mainly because their effects are somewhat overlapping: InLD weakens the feature response of the non-object region while weakening the image noise interference.

Therefore, ImLD is an optional module depending on the dataset and  computing environment. We will not use ImLD in subsequent experiments unless otherwise stated.
\begin{figure*}[!tb]
	\centering
	\subfigure[Small vehicle and large vehicle (HBB task).]{
			\centering
			\includegraphics[width=0.7\textwidth]{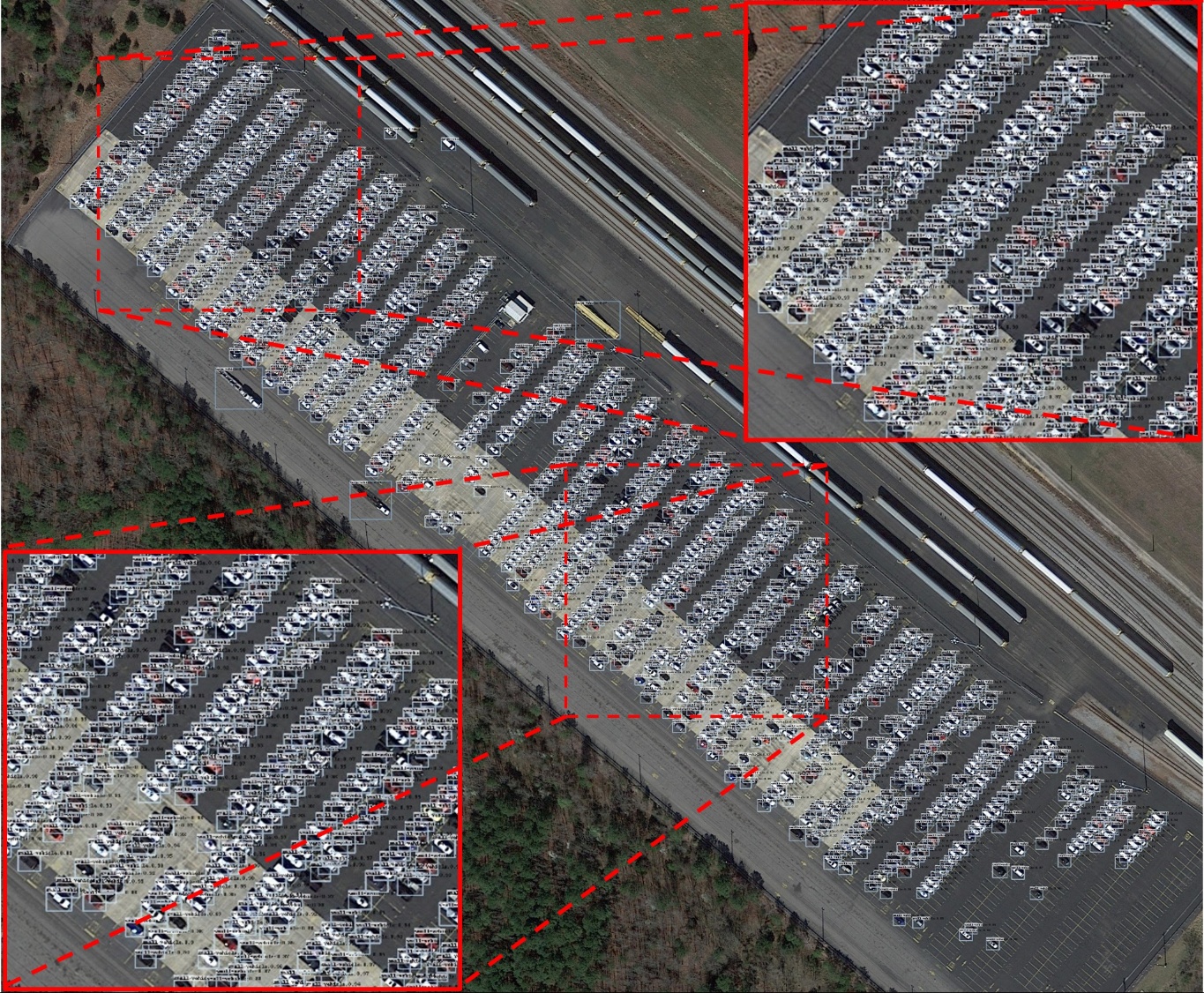}}\\
	\subfigure[Plane (OBB task).]{
			\centering
	\includegraphics[width=0.7\textwidth]{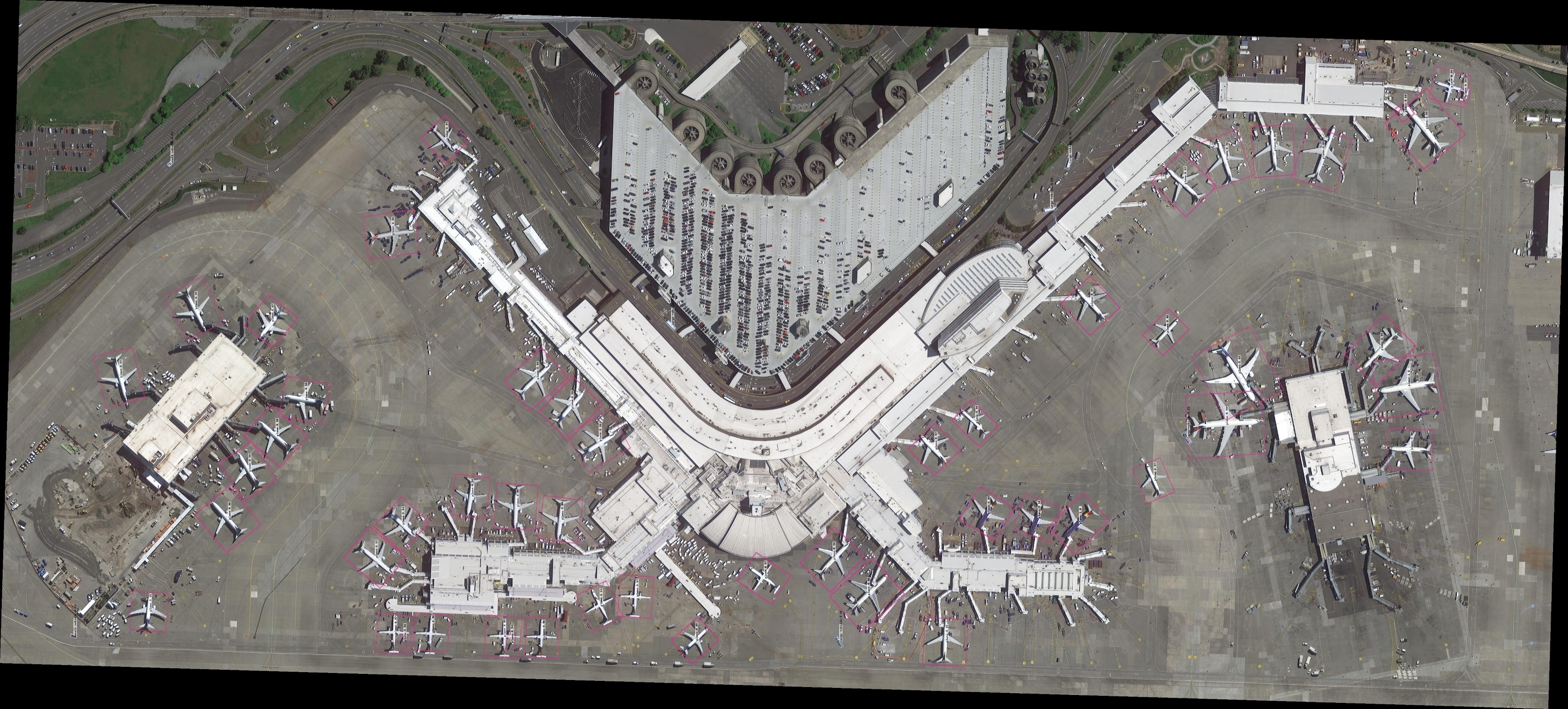}}
	\caption{Detection examples of our proposed method in large scenarios on DOTA-v1.0 dataset. Our method can both effectively handle the dense (top plot with white bounding box) and rotating (bottom plot with red bounding box) cases. Zoom in for better view.}
	\label{fig:LS}
	%\vspace{-5pt}
\end{figure*}

\textbf{Effect of IoU-Smooth L1 Loss on detectors and datasets.}
The IoU-Smooth L1 loss\footnote{Source code of IoU-Smooth L1 Loss is separately available at:  \url{https://github.com/yangxue0827/RotationDetection}} eliminates the boundary effects of the angle, making it easier for the model to regress to the objects coordinates. Tab. \ref{table:iou-smooth-l1} shows that new loss improves three detectors' accuracy to 69.83\%, 68.65\% and 76.20\%. Angle direct regression (Reg.) always suffer from boundary discontinuity. In contrast, angle indirect regression (Reg$^*$.) is a simpler way to avoid it and has an advantage in DOTA-v1.0, DOTA-v1.5 and DOTA-v2.0 according to Tab. \ref{tab:baseline}. IoU-Smooth L1 Loss further improves the performance to 66.99\%, 59.16\% and 46.31\% on three datasets.

\textbf{Effect of data augmentation and backbone.}
Using ResNet101 as backbone and data augmentation (random horizontal, vertical flipping, graying, and rotation), we observe a reasonable improvement in Tab. \ref{table:dota_ablation_study} (69.81\% $\rightarrow$ 72.98\%). We improve the final performance of the model from 72.98\% to 74.41\% by using ResNet152 as backbone. Due to the extreme imbalance of categories in the dataset, this provides a notable advantage to data augmentation, but we have found that this does not affect the functioning of InLD under these heave settings, from 72.81\% to 74.41\%. All experiments are performed on the OBB task on DOTA-v1.0, and the final model based on R$^3$Det is also named R$^3$Det++\footnote{Code of R$^3$Det and R$^3$Det++ are all available at \url{https://github.com/Thinklab-SJTU/R3Det_Tensorflow}.}.

\subsection{Comparison with the State-of-the-Art Methods}\label{subsec:sota}
We compare our proposed InLD with the state-of-the-art algorithms on two datasets DOTA-v1.0 \cite{xia2018dota} and DIOR \cite{li2020object}. Our model outperforms all other models.

\textbf{Results on DOTA-v1.0.}
We compare our results with the state-of-the-arts results in DOTA-v1.0 as depicted in Tab. \ref{table:dota_sota}. The results of DOTA-v1.0 reported here are obtained by submitting our predictions to the official DOTA-v1.0 evaluation server\footnote{\url{https://captain-whu.github.io/DOTA/evaluation.html}}. In the OBB task, we add the proposed InLD module to a single-stage detection method (R$^3$Det++) and a two-stage detection method (FPN-InLD). Our methods achieve the best performance, 76.56\% and 76.81\% respectively. To make fair comparison, we do not use overlays of various tricks, oversized backbones, and model ensemble, which are often used on DOTA's leaderboard methods. In the HBB task, we also conduct the same experiments and obtain competitive detection mAP, about 74.37\% and 76.24\%. Model performance can be further improved to 79.35\% if multi-scale training and testing are used. It is worth noting that FADet \cite{li2019feature}, SCRDet \cite{yang2019scrdet} and CAD-Det \cite{zhang2019cad} use the simple attention mechanism by Eq. \ref{eq:attention}, but our performance is far better than all. Fig. \ref{fig:DOTA} shows some aerial sub-images, and Fig. \ref{fig:LS} shows the aerial images of large scenes. In general, our method has the following two advantages over other methods:
i) we have solved the boundary problem in rotation detection, which is not considered by many methods; ii) an instance level denoising method is used, which is very helpful for complex remote sensing images.

\begin{table}[tb!]
	\centering
	\caption{Performance by accuracy (\%) on UCAS-AOD dataset.}
	\resizebox{0.3\textwidth}{!}{
		\begin{tabular}{l|c|c|c}
			\hline
			Method & mAP &  Plane & Car \\
			\hline
			YOLOv2 \cite{redmon2017yolo9000} & 87.90 & 96.60 & 79.20 \\
			R-DFPN \cite{yang2018automatic} & 89.20 & 95.90 & 82.50 \\
			DRBox \cite{liu2017learning} & 89.95 & 94.90 & 85.00 \\ 
			S$^2$ARN \cite{bao2019single} & 94.90 & 97.60 & 92.20 \\
			RetinaNet-H \cite{ yang2021r3det} & 95.47 & 97.34 & 93.60 \\
			ICN \cite{azimi2018towards} & 95.67 & -- & -- \\
			FADet \cite{li2019feature} & 95.71 & 98.69 & 92.72 \\
			R$^3$Det \cite{ yang2021r3det} & 96.17 & 98.20 & 94.14 \\
			\hline
			SCRDet++ (R$^3$Det)  & \textbf{96.95} & \textbf{98.93} & \textbf{94.97} \\
			\hline
	\end{tabular}}
	\label{table:UCAS-AOD}
	\vspace{-5pt}
\end{table}

\textbf{Results on DIOR and UCAS-AOD.}
DIOR is a new large-scale aerial images dataset, and has more categories than DOTA. In addition to the official baselines, we also give our final detection results in Tab. \ref{table:dior_sota}. It should be noted that the baseline we reproduce is higher than the official one. In the end, we obtain 77.80\% and 75.11\% mAP on FPN and RetinaNet based methods. Tab. \ref{table:UCAS-AOD} illustrates the comparison of performance on UCAS-AOD dataset. As we can see, our method achieves 96.95\% for OBB task and is the best out of all the existing published methods.

\section{Conclusion}\label{sec:conclusion}
We have presented an instance level denoising technique in the feature map for improving detection especially for small and densely arranged objects e.g. in aerial images. The core idea of InLD is to make the feature of different categories decoupled over different channels, while the features of the object and non-object are enhanced and weakened in the space, respectively. Meanwhile, the IoU constant factor is added to the smooth L1 loss to address the boundary problem in rotation detection for more accurate rotation estimation. We perform extensive ablation studies and comparative experiments on multiple aerial image datasets such as DOTA, DIOR, UCAS-AOD, small traffic light dataset BSTLD and our released S$^2$TLD, and demonstrate that our method achieves the state-of-the-art detection accuracy. We also use natural image dataset COCO and scene text dataset ICDAR2015 to verify the effectiveness of our approach. 

%\appendices
%% you can choose not to have a title for an appendix
%% if you want by leaving the argument blank
%\section{}

\section*{Acknowledgment}
This work was supported by National Key Research and Development Program of China (2020AAA0107600), and NSFC (U20B2068, U19B2035), Shanghai Municipal Science and Technology Major Project (2021SHZDZX0102). The author Xue Yang is supported by Wu Wen Jun Honorary Doctoral Scholarship, AI Institute, Shanghai Jiao Tong University. 

\ifCLASSOPTIONcaptionsoff
  \newpage
\fi

\bibliographystyle{IEEEtran}
\bibliography{IEEEegbib}

\begin{IEEEbiography}[{\includegraphics[width=1in,height=1.25in,clip,keepaspectratio]{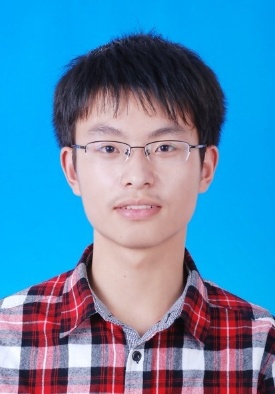}}]{Xue Yang} is currently a Ph.D. candidate with Department of Computer Science and Engineering, Shanghai Jiao Tong University, Shanghai, China. He received the B. E. degree from School of Information Science and Engineering, Central South University, Hunan, China, in 2016. He also received the M. S. degree from School of Electronic, Electrical and Communication Engineering, Chinese Academy of Sciences University, Beijing, China, in 2019. His research interests mainly include computer vision and machine learning, especially for object detection. He has published first-authored papers in IJCV, CVPR, ECCV, ICCV, ICML, NeurIPS, AAAI during his Ph.D. study from 2019 to 2022, and served as reviewers for NeurIPS, ICML, CVPR, ECCV, AAAI, ACM MM, IJCV, IEEE TIP etc. His Github projects on object detection have received over 5,000 stars, and ported to the projects MMRotate and AlphaRotate.

\end{IEEEbiography}
\begin{IEEEbiography}[{\includegraphics[width=1in,height=1.25in,clip,keepaspectratio]{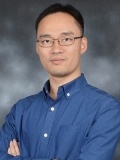}}]{Junchi Yan}(S'10-M'11-SM'21) is currently an Associate Professor with Department of Computer Science and Engineering, and AI Institute of Shanghai Jiao Tong University. Before that, he was a Senior Research Staff Member with IBM Research where he started his career since April 2011, and obtained his PhD from Electrical Engineering, Shanghai Jiao Tong University in 2015. His research interests include machine learning and computer vision. He serves Senior PC for CIKM 2019, IJCAI 2021, Area Chair for ICPR 2020, CVPR 2021, ACM-Multimedia 2021/2022, AAAI 2022, ICML 2022, Associate Editor for Pattern Recognition and IEEE ACCESS.
\end{IEEEbiography}

\begin{IEEEbiography}[{\includegraphics[width=1in,height=1.25in,clip,keepaspectratio]{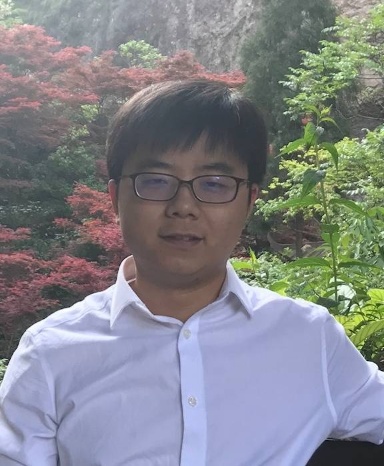}}]{Wenlong Liao} received the B. E. degree from Northwestern Polytechnical University, in the major of Detection, Guidance and Control Techniques in 2011, and M. S. degree from Shanghai Jiao Tong University in Control Science and Engineering in 2014. Since then he has been working on autonomous driving and currently a PhD candidate with Department of Computer Science and Engineering, Shanghai Jiao Tong University, with research interests in robotics.
\end{IEEEbiography}

\begin{IEEEbiography}[{\includegraphics[width=1in,height=1.25in,clip,keepaspectratio]{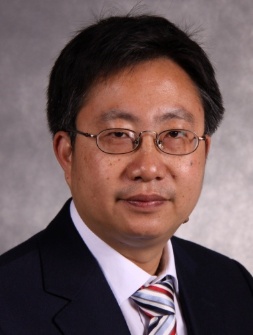}}]{Xiaokang Yang} (M'00-SM'04-F'19) received the B. S. degree from Xiamen University, in 1994, the M. S. degree from Chinese Academy of Sciences in 1997, and the Ph.D. degree from Shanghai Jiao Tong University in 2000. He is currently a Distinguished Professor, Shanghai Jiao Tong University, Shanghai, China. His research interests include visual signal processing and pattern recognition. He serves as an Associate Editor of IEEE Transactions on Multimedia.
\end{IEEEbiography}

\begin{IEEEbiography}[{\includegraphics[width=1in,height=1.25in,clip,keepaspectratio]{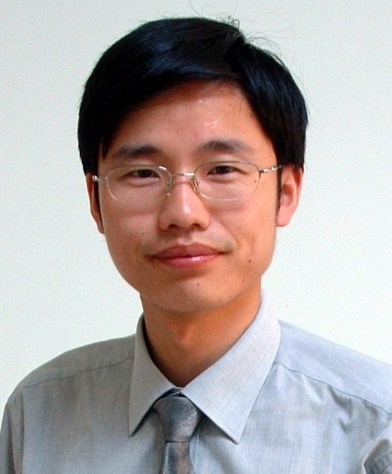}}]{Jin Tang} received the B.Eng. degree in automation and the Ph.D. degree in computer science from Anhui University, Hefei, China, in 1999 and 2007, respectively. He is currently a Professor with the School of Computer Science and Technology, Anhui University, Hefei, China. His current research interests include computer vision, pattern recognition, and deep learning.
\end{IEEEbiography}
\begin{IEEEbiography}[{\includegraphics[width=1in,height=1.25in,clip,keepaspectratio]{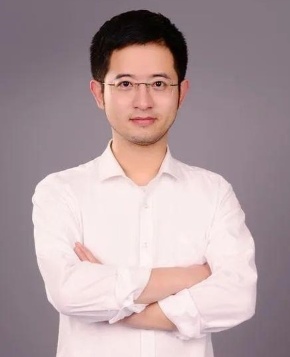}}]{Tao He} received the B. E. and M. E. degrees from Shanghai Jiao Tong University in Electrical Engineering in 2005 and 2008, respectively. He received his PhD in Mechanical and Aerospace Engineering, from Tokyo Institute of Technology, Tokyo, Japan in 2012. Since then he has been working on autonomous driving for and currently he is the founder and the Chief Executive Officer (CEO) of COWAROBOT Co., Ltd. He is the winner of Forbes 40 under 40 China.
\end{IEEEbiography}

\end{document}